\documentclass[runningheads]{llncs}

\usepackage{eccv}

\usepackage{eccvabbrv}

\usepackage{graphicx}
\usepackage{booktabs}

\usepackage[accsupp]{axessibility}  

\usepackage[pagebackref,breaklinks,colorlinks,citecolor=eccvblue]{hyperref}

\usepackage{orcidlink}
\usepackage{wrapfig}
\usepackage{graphicx}

\usepackage{subcaption}
\usepackage{amsmath,amssymb} 
\usepackage{color}
\usepackage{algorithm}
\usepackage{xcolor}
\usepackage{bm}
\usepackage{booktabs}
\usepackage[accsupp]{axessibility}
\usepackage{epsfig}
\usepackage{epigraph}
\usepackage[font={small}]{caption}
\usepackage{threeparttable}
\usepackage{tabularx}
\usepackage{tabularray}
\usepackage{wrapfig}
\usepackage{multirow}
\usepackage{float}
\usepackage{overpic}
\usepackage{booktabs} 
\usepackage{diagbox}
\usepackage{makecell}
\usepackage{bbm}
\usepackage{threeparttable}
\usepackage{dsfont}
\usepackage{arydshln}
\usepackage[misc]{ifsym}
\usepackage{makecell}
\usepackage{algpseudocode}

\def\rvx{{\mathbf{x}}}
\def\rvz{{\mathbf{z}}}
\def\rvc{{\mathbf{c}}}
\newcommand{\E}{\mathbb{E}}

\newcommand{\R}{\mathbb{R}}
\definecolor{top1}{RGB}{60,70,190}
\definecolor{top2}{RGB}{240,123,63}
\definecolor{top3}{RGB}{150,97,127}
\definecolor{changed}{RGB}{255,0,0}

\newcommand{\redlink}[1]{{\color{red}#1}}

\begin{document}

\title{EMDM: Efficient Motion Diffusion Model for Fast and High-Quality Motion Generation} 

\titlerunning{EMDM}

\author{Wenyang Zhou\inst{1}\orcidlink{0009-0006-0575-0523} \and
Zhiyang Dou\inst{2,3,4}$^{,\dag,}$\thanks{Collaborating with Tencent Games.}\orcidlink{0000-0003-0186-8269}\and Zeyu Cao\inst{1} \orcidlink{0000-0001-8269-3602} \and Zhouyingcheng Liao\inst{2}\orcidlink{0009-0002-6525-1372} \and Jingbo Wang\inst{5}\orcidlink{0009-0005-0740-8548} \and Wenjia Wang\inst{2}\orcidlink{0000-0003-0121-3852} \and Yuan Liu\inst{2}\orcidlink{0000-0003-2933-5667} \and Taku Komura\inst{2}\orcidlink{0000-0002-2729-5860} \and \\ Wenping Wang\inst{6}\orcidlink{0000-0002-2284-3952} \and Lingjie Liu\inst{3}\orcidlink{0000-0003-4301-1474}}

\authorrunning{Zhou et al.}

\institute{University of Cambridge \and The University of Hong Kong \and University of Pennsylvania \and TransGP \and Shanghai AI Laboratory  \and Texas A\&M University\\
\footnotesize \textit{$^{\dag}$Project Lead.}}

\maketitle

\begin{abstract}
We introduce Efficient Motion Diffusion Model (EMDM) for fast and high-quality human motion generation. Current state-of-the-art generative diffusion models have produced impressive results but struggle to achieve fast generation without sacrificing quality. On the one hand, previous works, like motion latent diffusion, conduct diffusion within a latent space for efficiency, but learning such a latent space can be a non-trivial effort. On the other hand, accelerating generation by naively increasing the sampling step size, e.g., DDIM, often leads to quality degradation as it fails to approximate the complex denoising distribution. To address these issues, we propose EMDM, which captures the complex distribution during multiple sampling steps in the diffusion model, allowing for much fewer sampling steps and significant acceleration in generation. This is achieved by a conditional denoising diffusion GAN to capture multimodal data distributions among arbitrary (and potentially larger) step sizes conditioned on control signals, enabling fewer-step motion sampling with high fidelity and diversity. To minimize undesired motion artifacts, geometric losses are imposed during network learning. As a result, EMDM achieves real-time motion generation and significantly improves the efficiency of motion diffusion models compared to existing methods while achieving high-quality motion generation. Our code is available at \url{https://github.com/Frank-ZY-Dou/EMDM}.

\end{abstract}

\section{Introduction}
\label{sec:intro}
\begin{figure}
\centering
\includegraphics[width=0.95\linewidth]{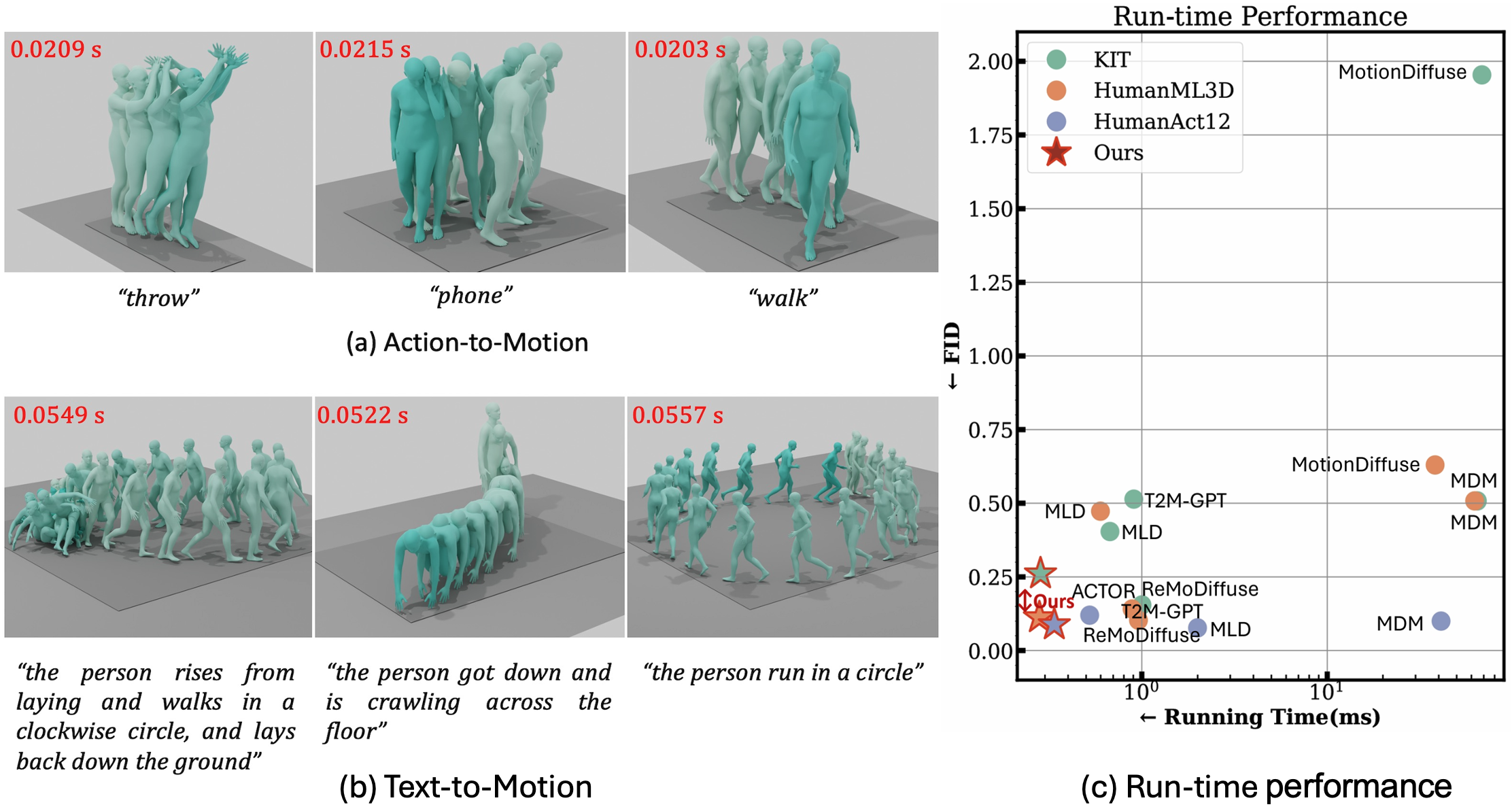} 
\caption{EMDM produces high-quality human motion aligned with input conditions in a short runtime. The average run time of EMDM in (a) action-to-motion and (b) text-to-motion tasks is $0.02$s and $0.05$s per sequence, respectively. For reference, the corresponding times for MDM~\cite{mdm2022human} are $2.5$s and $12.3$s. We deepen the color of the character with respect to the time step of the sequence. (c) Overall comparison of the inference time costs on the HumanML3D, KIT, and HumanAct12 datasets. For ease of illustration, the Running Time is plotted with a log scale. We compare the running time per frame vs. the FID of SOTA methods.}
\label{fig:teasar}
\end{figure}

Tremendous efforts have been made for human motion generation with different modalities, including action labels~\cite{petrovich21actor, guo2020action2motion, dou2023c, lee2023multiact,xu2023actformer}, textual descriptions~\cite{zhang2023tapmo, dabral2023mofusion, zhang2023t2m, kong2023priority, Guo_2022_CVPR_humanml3d,petrovich22temos, zhang2023remodiffuse, mdm2022human,chuan2022tm2t,ahuja2019language2pose,kim2022flame,jiang2024motiongpt}, and audio~\cite{li2021ai, li2022danceformer, lee2019dancing,alexanderson2023listen}, etc. The diffusion model~\cite{ho2020denoising,song2020denoising,po2023state} has been at the forefront of these advances~\cite{mdm2022human, chen2023executing, zhang2022motiondiffuse, shi2023controllable, karunratanakul2023guided}, due to its promise of effectively capturing the target distribution of diverse body motions. However, these models struggle to achieve fast motion generation while maintaining high motion quality. For instance, MDM~\cite{mdm2022human} takes around $12$s to produce a motion sequence given a textual description. Such low efficiency limits their effectiveness in real-world applications, e.g., online motion synthesis.

\begin{wrapfigure}[13]{r}{4.7cm}
  \begin{center}
\includegraphics[width=\linewidth]{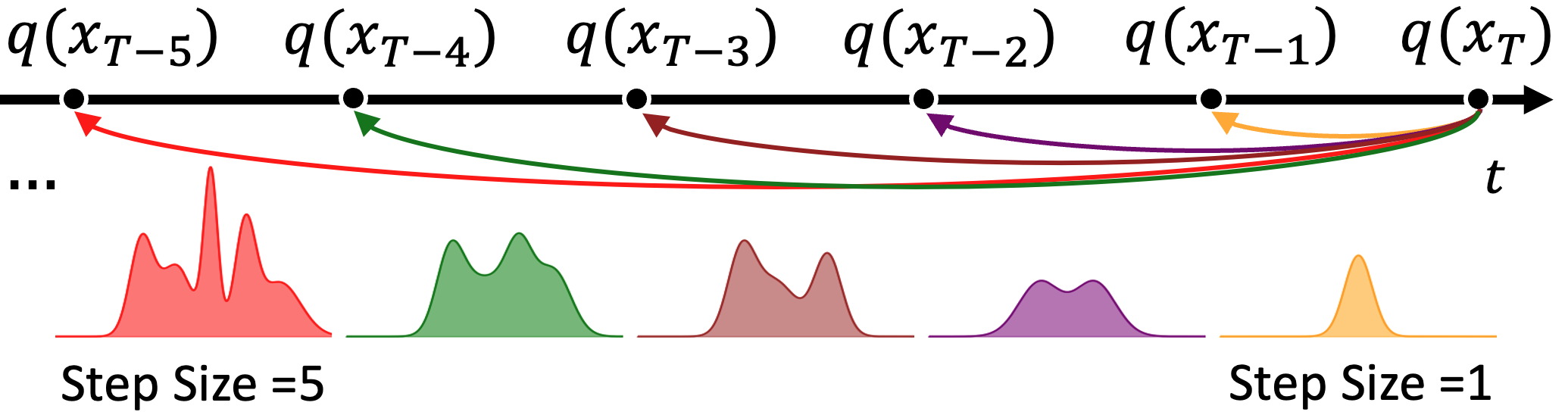}
  \end{center}
  \caption{Denoising distribution becomes complex (non-Gaussian) when increasing sampling step sizes for few-step sampling.}
  \label{fig:dist}
\end{wrapfigure}

Existing efforts to improve the generation efficiency of the motion diffusion model can be mainly categorized into two types: 1) motion latent diffusion proposed by MLD~\cite{chen2023executing}. This involves first learning a latent space of body motion and then conducting latent diffusion. However, such a two-stage approach relies on effectively embedding the motion in the first stage—it is challenging to learn a good embedding space for the subsequent latent diffusion model. The expression of the latent space typically limits the performance of downstream motion generation, as evidenced by both quantitative (Sec.~\ref{sec:comp:text} and Sec.~\ref{sec:comp:action}) and qualitative comparisons. 2) The DDIM sampling strategy~\cite{song2020denoising} can be adopted to accelerate generation by reducing the number of denoising steps (using a larger step size), given that the standard number of denoising steps is typically $1000$~\cite{mdm2022human,zhang2022motiondiffuse}. Additionally, the Gaussian assumption on the denoising distribution holds only for small step sizes. Therefore, directly using a larger step size during motion sampling skips numerous reverse steps and leads to much more complex data distributions than Gaussians. As the complex data distributions cannot be approximated with fewer sampling steps, the performance of this approach drops, as shown in Tab.~\redlink{D2} in Appendix~\redlink{D}. Thus, it is critical to capture the complex data distributions when a few-step sampling is involved; see Fig.~\ref{fig:dist}.

\begin{figure}[t]
\centering
\begin{overpic}
[width=0.95\linewidth]{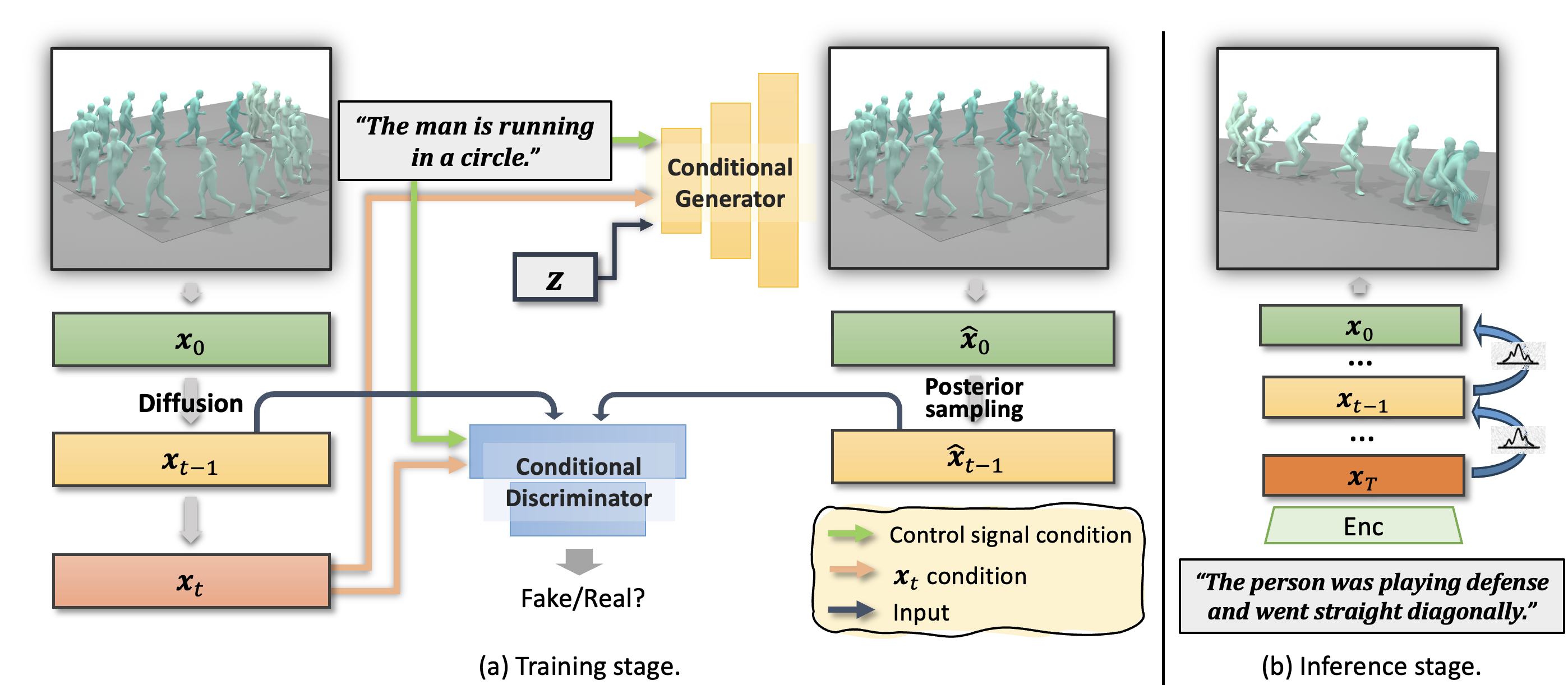} 
\end{overpic}
\caption{Pipeline of EMDM. We develop condition denoising diffusion GAN to capture the complex denoising distribution of human body motion, allowing a larger sampling step size (Sec.~\ref{sec:emdm}). During inference, we use a larger sampling step for fast sampling of high-quality motion w.r.t. input condition. The detailed sampling algorithm is given in Alg.~\ref{alg:sample_from_model}. Note that we ignore the time step $t$ for simplicity.}
\label{fig:pipeline}
\end{figure}

In this paper, we present Efficient Motion Diffusion Model (EMDM) for fast and high-quality human motion generation. We seek to reduce the number of sampling steps while achieving fast motion generation. The key to allowing a larger sampling step size is to effectively capture the complex data distributions during a few-step sampling. Inspired by recent advances in efficient image synthesis~\cite{xiao2021tackling}, we develop a sampling strategy for fast motion generation while maintaining high motion quality. Specifically, we employ a conditional denoising diffusion GAN, incorporating a conditional generator and conditional discriminator that consider both the time step $t$ and input control signals (e.g., text); See Fig.~\ref{fig:pipeline}. The generator (denoiser) is trained to generate the motion $\hat x_0$ conditioning on input control signals, time step $t$, given the random variables. Then posterior sampling~(Alg.~\ref{alg:sample_posterior}) is applied to produce $\hat x_{t-1}$ at the $t-1$-th time step using $\hat x_0$. A discriminator is trained to distinguish whether a data sample $\hat x_{t-1}$ is a plausible denoised result of $x_t$. As $t$ varies during diffusion model training, the generator learns to capture the complex denoising distribution introduced by an arbitrary (and potentially larger) sampling step size.  As a result, during sampling, one could use a larger sampling step size (fewer steps) to sample a motion given the conditions, significantly improving runtime performance. As a condition of the model, the control signals make the capture of the complex motion distribution more efficient by learning the conditional denoising distribution. Finally, to reduce unwanted artifacts, we further integrate geometric motion losses during model training to stabilize the training process and enhance motion quality. Our model is trained end-to-end, simplifying the training process and significantly reducing the overall training effort, which is a noteworthy advantage in practical applications.

As a result, EMDM effectively captures the complex motion distribution, enabling much fewer sampling steps during motion generation while maintaining high-quality motion; See Fig.~\ref{fig:teasar} for some examples of generated motion and overall running time statistics. Our contribution is three-fold:
\begin{itemize}
 \item We reveal the efficiency issues with existing motion diffusion models and the challenges in accelerating the models.
 \item We present EMDM for fast and high-quality motion generation by employing a conditional denoising diffusion GAN to effectively model complex denoising distributions for the few-step motion generation with high quality.
 \item 
 We perform extensive experiments on EMDM to demonstrate its remarkable speed-up for diffusion-based approaches with competitive or even higher quality and diversity of the generated motions compared with SOTAs.
 \end{itemize}

\section{Related Work}
\label{sec:related}
\textbf{Human Motion Generation} 
Human motion generation is an important research problem in computer vision and computer animation. The ability to generate realistic and natural human motions has wide applications including virtual reality~\cite{questsim,questenvsim,neural3points}, game development~\cite{tessler2023calm, starke2019neural, starke2021neural,  holden2017phase, starke2022deepphase}, human behavior analysis~\cite{liu2022close, guo2023student,zhang2023popularization,  zhang2023close, yang2023analysis,chong2020detection,crawford2022impact} and robotics~\cite{wan2022learn, smith2023learning,li2023robust,yamane2013synthesizing,christen2023learning}. The generated motion can condition on abundant, multi-modal inputs such as action labels~\cite{petrovich21actor, guo2020action2motion, dou2023c, lee2023multiact,xu2023actformer}, textual description~\cite{chen2024taming, voas2309best, zhang2023tapmo, zhang2023t2m, dabral2023mofusion, Guo_2022_CVPR_humanml3d, mdm2022human, chuan2022tm2t, ahuja2019language2pose, kong2023priority, zhang2023remodiffuse, petrovich2022temos, kim2022flame,ao2023gesturediffuclip, zhang2023motiongpt},  incomplete pose sequences~\cite{wan2022,duan2021single, harvey2020robust, mdm2022human}, control signals~\cite{li2023aamdm,starke2019neural, 2021-TOG-AMP, xie2023omnicontrol, shi2023phasemp, holden2017phase,starke2022deepphase, pi2023hierarchical, liao2022skeleton, wan2023diffusionphase, cong2024laserhuman, wan2023tlcontrol, zhang2023skinned}, music or audio~\cite{li2021ai, li2022danceformer, lee2019dancing, pang2023bodyformer,ao2022rhythmic, zhu2023taming}, and so on. For \textit{Unconditional Motion Generation}~\cite{yan2019convolutional,zhao2020bayesian,zhang2020perpetual, raab2022modi, mdm2022human, rempe2021humor}, the goal is to model the entire motion space based on motion data. For instance, VPoser~\cite{vposer_SMPL-X:2019} introduces a variational human pose prior primarily for image-based pose fitting, while ACTOR~\cite{petrovich21actor, petrovich22temos} presents a class-agnostic transformer VAE as a baseline. Humor~\cite{rempe2021humor} employs a conditional VAE for learning motion prior in an auto-regressive manner. The recent study~\cite{shi2023phasemp} learns phase-conditioned motion prior in the frequency domain. \textit{Action-to-Motion}~\cite{guo2020action2motion, dou2023c, lee2023multiact,xu2023actformer, petrovich21actor} can be viewed as the inverse task of the classical action recognition task, where the goal is to produce human motion given the input action labels. Specifically, ACTOR~\cite{petrovich21actor} introduces learnable biases within a transformer VAE to encapsulate action for motion generation. Nowadays, \textit{Text-to-Motion}~\cite{petrovich22temos, ahuja2019language2pose, zhang2022motiondiffuse, mdm2022human, kim2022flame, Guo_2022_CVPR_humanml3d,kong2023priority, zhang2023remodiffuse,ao2023gesturediffuclip, zhang2023t2m} has become popular, primarily because of the user-friendly and accessible nature of language descriptors. Specifically, T2M-GPT~\cite{zhang2023t2m} proposes a classic framework based on VQ-VAE and GPT to synthesize human motion from textual descriptions. \cite{mdm2022human, zhang2022motiondiffuse} employ diffusion models for high quality text-to-motion. Recently,~\cite{jiang2024motiongpt, zhang2023motiongpt} propose motion language pre-training using LLMs~\cite{touvron2023llama, raffel2020exploring, chung2022scaling} for text-driven motion synthesis.\\
\noindent\textbf{Motion Diffusion Models.} Diffusion Generative Models~\cite{sohl2015deep} have shown impressive results in wide fields~\cite{po2023state, stable_diffusion, muller2023diffrf, ho2022imagen, liu2023syncdreamer, long2023wonder3d, chou2023diffusion, yu2023surf, zhang2022motiondiffuse, mdm2022human} and Diffusion models have been employed for human motion generation~\cite{mdm2022human, zhang2022motiondiffuse, kim2022flame,chen2023executing, shi2023controllable, xie2023omnicontrol, karunratanakul2023guided}. Specifically, MotionDiffuse~\cite{zhang2022motiondiffuse} stands as the first text-based motion diffusion model using fine-grained instructions for body part-level control. MDM~\cite{mdm2022human} conducts motion diffusion that operates on raw motion data, learning the relationship between motion and input conditions. 
ReMoDiffuse~\cite{zhang2023remodiffuse} presents a retrieval-augmented motion diffusion model, where extra knowledge from the retrieved samples is used for motion synthesis. Recent efforts~\cite{karunratanakul2023guided, xie2023omnicontrol} have concentrated on controllable human motion generation, leveraging either pelvis location~\cite{karunratanakul2023guided} or specific body joints~\cite{xie2023omnicontrol}. That being said, applying the diffusion model to the motion data~\cite{mdm2022human, zhang2022motiondiffuse} as a sequential motion generation framework incurs high computational overheads and typically results in low inference speeds due to model size and their iterative sampling nature. To tackle the problem, MLD~\cite{chen2023executing} introduces a motion latent-based diffusion model by first training a VAE for motion embedding, followed by the application of latent diffusion within the learned latent space. However, this is a two-stage method and requires non-end-to-end training: effectively capturing the motion distribution during motion embedding can be challenging, yet it is crucial for the success of the second stage. In contrast, our approach aims to boost efficiency by accelerating the sampling process and is end-to-end trainable. Retrieval-based method~\cite{zhang2023remodiffuse} could achieve relatively fast motion generation. As of yet, it relies on reference motion datasets and suffers from relatively low motion diversity. EMDM allows for much fewer sampling steps during the denoising process without the reliance on the reference motion for sequential motion generation with high quality.

\section{Method}
\label{sec:method}
Our goal is to efficiently generate high-quality and diverse human motion given conditional inputs in {real time}. We propose an Efficient Motion Diffusion Model utilizing a conditional denoising diffusion GAN for fast motion generation, which will be elaborated in the following. 

\subsection{Efficient Motion Diffusion Model}
\label{sec:emdm}
In this task, the motion of humans, denoted as $\rvx^{1:N}$, is associated with a corresponding condition $\rvc$, e.g., action~\cite{guo2020action2motion,petrovich21actor, mdm2022human} or text~\cite{tevet2022motionclip,Guo_2022_CVPR_humanml3d, zhang2022motiondiffuse, mdm2022human}. $N$ is the number of frames in a motion sequence. Note that unconditioned motion generation is available by $\rvc=\emptyset$ similar to~\cite{mdm2022human, chen2023executing}. We use probabilistic diffusion models~\cite{sohl2015deep} for motion generation. The forward process of the diffusion model is given by
\begin{equation}
{
q(\rvx^{1:N}_{1:T}|\rvx^{1:N}_0) = \prod_{t \ge 1} q(\rvx^{1:N}_t|\rvx^{1:N}_{t-1}), \quad q(\rvx^{1:N}_{t} | \rvx^{1:N}_{t-1}) = \mathcal{N}(\sqrt{\alpha_{t}}\rvx^{1:N}_{t-1},(1-\alpha_{t})\mathbf{I}),
}
\label{eq:forward}
\end{equation} 
where $\alpha_{t} \in (0,1)$ are constant hyper-parameters. When $\alpha_{t}$ is small enough, we can approximate $\rvx^{1:N}_{T} \sim \mathcal{N}(0,\mathbf{I})$~\cite{sohl2015deep}. The reverse process is given by
\begin{align}
\centering
     p_{\theta}(\rvx^{1:N}_{0:T}) = p(\rvx^{1:N}_{T})\prod_{t \ge 1} p_{\theta}(\rvx^{1:N}_{t-1}|\rvx^{1:N}_{t}), p_\theta(\rvx^{1:N}_{t-1} | \rvx^{1:N}_{t}) = \mathcal{N}(\rvx^{1:N}_{t-1}; \bm{\mu}_\theta(\rvx^{1:N}_t, t), \sigma_t^2\mathbf{I}),
\label{eq:reverse}
\end{align} 
where $\theta$ is the learnable parameters of the diffusion model which gradually anneals the noise from a Gaussian distribution to the data distribution.

When training a motion diffusion model, a denoiser $\epsilon_\theta\left(\rvx_t, t\right)$ learns to iteratively anneal the random noise to the motion sequence $\{{\hat{\rvx}_t^{1:N}}\}^{T}_{t=1}$, where the human pose ${\rvx}^i\in\R^{J\times D}$ at the $i$-th frame is represented by either joint rotations or positions, where $J$ is the number of joints and $D$ is the dimension of the joint representation. When $\alpha_t$ is large, the denoising distribution $q(\rvx_{t-1} | \rvx_t)$ and $q(\rvx_{t} | \rvx_{t-1})$ can be both regarded as Gaussian. With this assumption, diffusion models often have thousands of steps with a large $\alpha_t$, e.g., MDM~\cite{mdm2022human} and MotionDiffuse~\cite{zhang2022motiondiffuse} need $1000$ steps for denoising, leading to a rather slow motion generation process. Obviously, when the denoising step size is naively increased (fewer sampling steps), i.e. in the case of DDIM sampling, the distribution is non-Gaussian; there is thus no guarantee that the Gaussian assumption on the denoising distribution holds (see Fig.~\ref{fig:dist}). Consequently, the quality of generated motions drops.

Inspired by the recent progress~\cite{xiao2021tackling} in image generation, we propose to model the expressive multimodal denoising distribution with a larger step size $q(\rvx_{t-1}|\rvx_t)$ by conditioning on the control signals and time step $t$. The training process is formulated by matching $p_{\theta}(\rvx_{t-1}|\rvx_t)$ and $q(\rvx_{t-1}|\rvx_t)$ when each diffusion step has smaller $\alpha_t$, which allows $T$ to be small ($T\leq10$).

\paragraph{Conditional Generator.} As $p_\theta(\rvx_{t-1}|\rvx_t) := q(\rvx_{t-1}|\rvx_t,\rvx_0 = g_\theta(\rvx_\theta,t))$ ~\cite{ho2020denoising}, one can first predict $\rvx_0$ using the diffusion model $g_\theta(\rvx_\theta,t)$ and then sample $\rvx_{t-1}$ using the posterior distribution $q(\rvx_{t-1} | \rvx_t,\rvx_0)$~\cite{xiao2021tackling}. In this paper, we employ a conditional denoising diffusion GAN, which integrates a conditional generator and conditional discriminator, considering both the time step $t$ and input control signals $\rvc$, for example text. To achieve motion denoising, the $g_\theta$ is modeled by a conditional generator $G_{\theta}(\rvx_t, \rvz, \rvc, t)$ that outputs $\hat \rvx_0$, given $\rvx_t$, control signal $\rvc$ and an $L$-dimensional latent variable $\rvz \sim p(\rvz):=\mathcal{N}(\rvz; \mathbf{0}, \mathbf{I})$. Mathematically, with $G_{\theta}(\rvx_t, \rvz, \rvc, t)$, $p_{\theta}(\rvx_{t-1} | \rvx_{t})$ is obtained by
\begin{equation}
\begin{aligned}
    p_{\theta}(\rvx_{t-1}|\rvx_t) &= \int p_{\theta}(\rvx_0|\rvx_t) q(\rvx_{t-1}|\rvx_t, \rvx_0) d\rvx_0 \\
    &= \int p(\rvz) q(\rvx_{t-1}|\rvx_t, \rvx_0\!=\!G_{\theta}(\rvx_t, \rvz, \rvc, t)) d\rvz.
\end{aligned}
\label{eq:q_x_0_prediction}
\end{equation}
We further use posterior distribution $q(\rvx_{t-1}|\rvx_t, \rvx_0)$ to sample $\hat \rvx_{t-1}$ for discrimination based on the predicted $\hat \rvx_0$ in the following.

\paragraph{Conditional Discriminator.}  We employ a time step-dependent and control signal-conditioned discriminator as $D_{\phi}(\rvx_{t-1}, \rvx_t, \rvc, t)$. The $N$-dimensional $\rvx_{t-1}$, $\rvx_t$ are two inputs at time step $t-1$ and $t$, and $\rvc$ is the control signal such as textual descriptions. It is trained to distinguish whether $\rvx_{t-1}$ is a plausible denoised result of $\rvx_{t}$. The discriminator is trained by

\begin{equation}
\begin{aligned}
    \min_{\phi} \sum_{t \geq 1} \mathbb{E}_{q(\rvx_t)}\ [\mathbb{E}_{q(\rvx_{t-1} | \rvx_{t})} [ \mathop{\mathrm{F}} (-D_\phi(\rvx_{t-1}, \rvx_t, \rvc, t))] +\\
    \mathbb{E}_{p_{\theta}(\rvx_{t-1} | \rvx_{t})} [\mathop{\mathrm{F}} ( D_\phi(\rvx_{t-1}, \rvx_t, \rvc, t))]\!,
\end{aligned}  
\label{eq:disc_training}
\end{equation}

where $\mathrm{F}(\cdot)$ denotes the $\mathrm{softplus}(\cdot)$ function and fake samples from $p_{\theta}(\rvx_{t-1} | \rvx_{t})$ are contrasted against real samples from $q(\rvx_{t-1} | \rvx_{t})$. By using the identity $q(\rvx_t, \rvx_{t-1}) = \int d\rvx_0 q(\rvx_0) q(\rvx_t, \rvx_{t-1} | \rvx_0) = \int d\rvx_0 q(\rvx_0) q(\rvx_{t-1} | \rvx_0) q(\rvx_{t}|\rvx_{t-1})$, we have 
\begin{equation}
\begin{aligned}
    \min_{\phi} \sum_{t \geq 1} (\E_{q(\rvx_0)q(\rvx_{t-1}|\rvx_0)q(\rvx_t | \rvx_{t-1})} [ \mathop{\mathrm{F}} (-D_\phi(\rvx_{t-1}, \rvx_t, \rvc, t))] \\
    +\mathbb{E}_{q(\rvx_t)}\mathbb{E}_{p_{\theta}(\rvx_{t-1} | \rvx_{t})} [\mathop{\mathrm{F}} ( D_\phi(\rvx_{t-1}, \rvx_t, \rvc, t))] ]).
\end{aligned}  
\label{eq:disc_training}
\end{equation}

Given the training goal of the condition discriminator in Eq.~\ref{eq:disc_training}, 
we can train the conditional generator $G_{\theta}(\rvx_t, \rvz, \rvc, t)$ by $\max_{\theta}\mathcal{L}_\text{disc}$, where $\mathcal{L}_\text{disc}$ is defined by 
\begin{equation}
\begin{aligned}
\mathcal{L}_\text{disc} = \mathbb{E}_{t \sim [1,T], q(\rvx_t)}\E_{p_{\theta}(\rvx_{t\!-\!1} | \rvx_{t})} [\mathop{\mathrm{F}} ( -D_\phi(\rvx_{t-1}, \rvx_t, \rvc, t))].
\end{aligned}  
\label{eq:disc_loss_func}
\end{equation}
The overall pipeline is shown in Fig.~\ref{fig:pipeline}. Our method can be taken for conditional generation, where the condition of the control signal (text) provides a strong clue for capturing the complex data distribution, thus effectively enhancing the overall model's performance compared with naively applying model~\cite{xiao2021tackling}, as shown in Sec.~\ref{sec:ablation}. After training, the conditional generator is used to sample motion with a few denoising steps, which we discuss in Sec.~\ref{sec:motion_sampling}.

\paragraph{Geometric Loss Functions.} Moreover, during training, we found the training scheme of the conditional denoising diffusion GAN to be inefficient, resulting in low-quality human motion results (see comparisons in Sec.~\ref{sec:ablation_geometric_loss}). We deem 
this is because motion generation requires more detailed constraints specifically tailored for the motion generation task, which cannot be effectively provided solely by the discrimination loss (Eq.~\ref{eq:disc_loss_func}). We thus employ geometric losses~\cite{mdm2022human} in addition to discrimination loss during model training to enhance motion quality. Specifically, for generator (denoiser), we follow~\cite{mdm2022human, chen2023executing} and predict the denoised motion itself, i.e., $\hat{\rvx}_{0} = G(\rvx_t, \rvz, \rvc, t)$ with the following losses on reconstruction, joint positions, foot contact, and joint velocities:
\begin{align}
    \mathcal{L}_\text{recon} &= E_{\rvx_0 \sim q(\rvx_0|\rvc), t \sim [1,T]}[\| \rvx_0 - G_\theta(\rvx_t, \rvz, \rvc,t)\|_2^2], \\
    \mathcal{L}_\text{pos} &= \frac{1}{N} \sum_{i =1}^{N} \| FK(\rvx^i_0) - FK(\hat{\rvx}^i_0)\|_{2}^{2}, \\
    \mathcal{L}_\text{foot} &= \frac{1}{N-1} \sum_{i =1}^{N-1} \| (FK(\hat{\rvx}^{i+1}_0) - FK(\hat{\rvx}^i_0)) \cdot f_i\|_{2}^{2}, \\
    \mathcal{L}_\text{vel} &= \frac{1}{N-1} \sum_{i =1}^{N-1} \| (\rvx^{i+1}_0 - \rvx^i_0) - (\hat{\rvx}^{i+1}_0 - \hat{\rvx}^i_0)\|_{2}^{2}
\end{align}

The geometric loss is thus given by
\begin{equation}
{
\mathcal{L}_{\text{geo}} =\mathcal{L}_\text{recon} + \lambda(\mathcal{L}_\text{pos} + \mathcal{L}_\text{vel} +\mathcal{L}_\text{foot}).
}
\end{equation}
Here, $FK(\cdot)$ denotes the forward kinematics converting joint rotations into joint positions. $f_i \in \{0,1\}^J$ is the binary foot contact mask for each frame $i$. Note that we use $\lambda$ as a binary indicator variable; in this paper, we set $\lambda$ to $1$ and $0$ for action-to-motion and text-to-motion tasks, respectively. We further investigate the effectiveness of the geometric loss functions in Sec.~\ref{sec:ablation_geometric_loss}. Finally, we train the generator using the overall objective with a balancing term $R$:
\begin{equation}
\begin{aligned}
{
\min_{\theta} (\mathcal{L}_{\text{disc}} + R \cdot \mathcal{L}_{\text{geo}}).
}
\end{aligned}  
\label{eq:total_loss}
\end{equation}

\subsection{Motion Sampling}
\label{sec:motion_sampling}
\begin{figure}[t]
\footnotesize
    \centering
    \begin{minipage}{0.5\textwidth}
        \begin{algorithm}[H]
            \caption{Sample from Model}
            \label{alg:sample_from_model}
                    \footnotesize
            \begin{algorithmic}[1]
                \Function{SAMPLE}{$\rvc$}
                    \Statex $\rvx_{T} \gets \text{random noise}$ 
                    \Statex $T \gets \text{the number of time steps}$ 
                    \Statex $G \gets \text{generator model}$ 
                    \Statex $\rvc \gets \text{the label (text or action number)}$ 
                    
                    \State $\rvx_t \gets \rvx_{T}$ 
                    \For{$t \gets T-1$ \textbf{to} $0$}
                        \State $\rvz \gets \Call{randn}{\rvz_{dim}}$
                               \\
                        \Comment{dimension of $\rvz$ is 64 in our paper.}
                        \State $\rvx_0 \gets \Call{generator}{\rvx_t, t, \rvz, \rvc}$
                        \State $\rvx_t \gets \Call{sample\_posterior(Alg.~\ref{alg:sample_posterior})}{\rvx_0, \rvx_t, t}$
                    \EndFor
                    \State \Return $\rvx_t$
                \EndFunction
            \end{algorithmic}
        \end{algorithm}
    \end{minipage}
    \hfill
    \begin{minipage}{0.48\textwidth}
        \begin{algorithm}[H]
            \caption{Sample Posterior}
            \label{alg:sample_posterior}
            \begin{algorithmic}[1]
                \Function{sample\_posterior}{$\rvx_0$, $\rvx_t$, $t$}
                \Statex $\text{coef1} \gets \text{posterior coefficient 1}$ 
                \Statex $\text{coef2} \gets \text{posterior coefficient 2}$ 
                
                    \State $mean \gets \text{coef1}[t] \times \rvx_0 + \text{coef2}[t] \times \rvx_t$
                    \State $log\_var \gets \text{posterior\_log\_variance}[t]$
                    \State $noise \gets \Call{randn\_like}{\rvx_t}$
                    \State $m \gets 0 \text{ if } t = 0 \text{ else } 1$
                    \State \Return $mean + m \times \text{exp}(0.5 \times log\_var) \times noise$
                \EndFunction
            \end{algorithmic}
        \end{algorithm}
    \end{minipage}
\end{figure}
We adopt classifier-free guidance~\cite{ho2022classifier} in EMDM. Following~\cite{chen2023executing}, our generator $G$ learns both the conditioned and the unconditioned motion generation task by randomly setting $\rvc=\emptyset$ for $10\%$ of the samples, such that $G(\rvx_t, \rvz, t, \emptyset)$ approximates $p(\rvx_0)$. When sampling from $G$, we trade off diversity and fidelity by interpolating or even extrapolating the two variants using $s$:
\begin{equation}
\begin{aligned}
G_s(\rvx_t, \rvz, \rvc, t) = G(\rvx_t, \rvz, \emptyset, t) + s\cdot (G(\rvx_t, \rvz, \rvc, t) - G(\rvx_t, \rvz, \emptyset, t))
\end{aligned}
\end{equation}
Given an input condition $\rvc$ which can be a sentence $\boldsymbol {w}^{1:N}=\{w^i\}_{i=1}^{N}$, a action label $a$ from the predefined action categories set $a \in A$~\cite{petrovich21actor} or even a empty condition $c = \varnothing$~\cite{vposer_SMPL-X:2019, zhang2021we}, 
EMDM aims to generate a human motion $\hat{\rvx}^{1:N}=\{\hat{\rvx}^i\}_{i=1}^{N}$ in a non-deterministic way, where $N$ denotes the motion length or frame number. Note that for the text-to-motion task, we employ the motion representation in ~\cite{mdm2022human,zhang2022motiondiffuse, Guo_2022_CVPR_humanml3d, chen2023executing}: a combination of 3D joint rotations, positions, velocities, and foot contact.  The sampling algorithm is specified in Alg.~\ref{alg:sample_from_model}.

\section{Experiments}
\label{sec:experiments}
We conduct extensive experiments to evaluate our models on motion quality and model efficiency. We test our model on multiple datasets for different motion synthesis tasks, including text-to-motion (\cref{sec:comp:text}) and action-to-motion (\cref{sec:comp:action}), qualitatively and quantitatively. The comparison with other few-step sampling methods for efficient motion generation is given in Appendix~\redlink{D1}. We evaluate unconditional motion generation in Appendix~\redlink{B}. More qualitative results can be found in Appendix~\redlink{A} and the supplementary video. We also conduct comparisons with the original DDGAN~\cite{xiao2021tackling} in Appendix~\redlink{D2}. \\
\noindent
\textbf{Datasets.} We use the following datasets for training and evaluating EMDM.\\
\noindent 
\textit{HumanML3D}~\cite{Guo_2022_CVPR_humanml3d} has $14616$ sequences from AMASS~\cite{AMASS_ICCV2019} with $44970$ textual description. \\
\noindent
\textit{KIT}~\cite{Plappert2016kit} collects $3911$ motions with $6353$ descriptions.\\
\textit{HumanAct12}~\cite{guo2020action2motion} provides $1191$ motion sequences and 12 action categories.\\
We use HumanML3D and KIT for the text-to-motion task while adopting HumanAct12 for the action-to-motion task. Refer to Appendix~\redlink{B} for unconditional motion generation.

\noindent
\textbf{Metrics.} We use the following metrics for evaluation.\\
\noindent \textit{Motion quality.} We use Frechet Inception Distance (FID) as a principal metric to evaluate the feature distributions between the generated and real motions. The feature used is extracted following the previous approach in~\cite{Guo_2022_CVPR_humanml3d}. \\
\noindent \textit{Motion diversity.}  Motion Diversity (DIV) calculates variance through features, and MultiModality (MM) measures the diversity using the same condition.\\
\noindent \textit{Condition matching.} Following~\cite{Guo_2022_CVPR_humanml3d},  we compute
motion-retrieval precision (R Precision) and report the text and motion Top 1/2/3 matching accuracy, and multi-modal distance (MM Dist) is used to calculate the distance between motions and texts. For action-to-motion, we use the corresponding action recognition models~\cite{guo2020action2motion, petrovich21actor} to calculate Accuracy (ACC) for action categories.\\ 
\noindent \textit{Run-time Performance.} We present the running time of various methods in milliseconds per frame as a metric to assess the inference efficiency of the models.

\subsection{Implementation Details}
\label{sec:implementation}
We use a transformer-based denoiser $\epsilon_\theta$ consisting of 12 layers and 32 heads with skip connections by default. The conditional discriminator is a 7-layer MLP network. The detailed architectures are given in Appendix~\redlink{C}. We employ a frozen \textit{CLIP-ViT-L-14}~\cite{radford2021learning} model as the text encoder $\tau_\theta^{w}$ for the text condition and a learnable embedding for action condition. All models are trained with the AdamW optimizer using a fixed learning rate of $3 \times 10^{-5}$ and $2 \times 10^{-5}$ for action-to-motion and text-to-motion, respectively. We use EMA decay on the optimizer during training. Our batch size is $64$ during the training stage. The model is trained on an Nvidia RTX 4090 GPU with an AMD 16-core CPU. For inference, we use an RTX 4090 GPU with an Intel 8-core CPU to run all experiments under identical settings for ten passes.

\subsection{Inference Time Costs}
We first present the overall comparison of inference time cost on both text-to-motion and action-to-motion tasks. As demonstrated in Fig.~\ref{fig:teasar}~(c), on both tasks, EMDM demonstrates the best or second-best performance in FID, simultaneously achieving superior efficiency in motion generation. Notably, although MLD~\cite{chen2023executing} and ReMoDiffuse~\cite{zhang2023remodiffuse} achieve competitive efficiency, these are two-stage methods that are non-end-to-end trainable. In contrast, EMDM exhibits a competitive or even better performance with reduced running time.

\begin{figure}[t]
    \centering
\includegraphics[width=\linewidth]{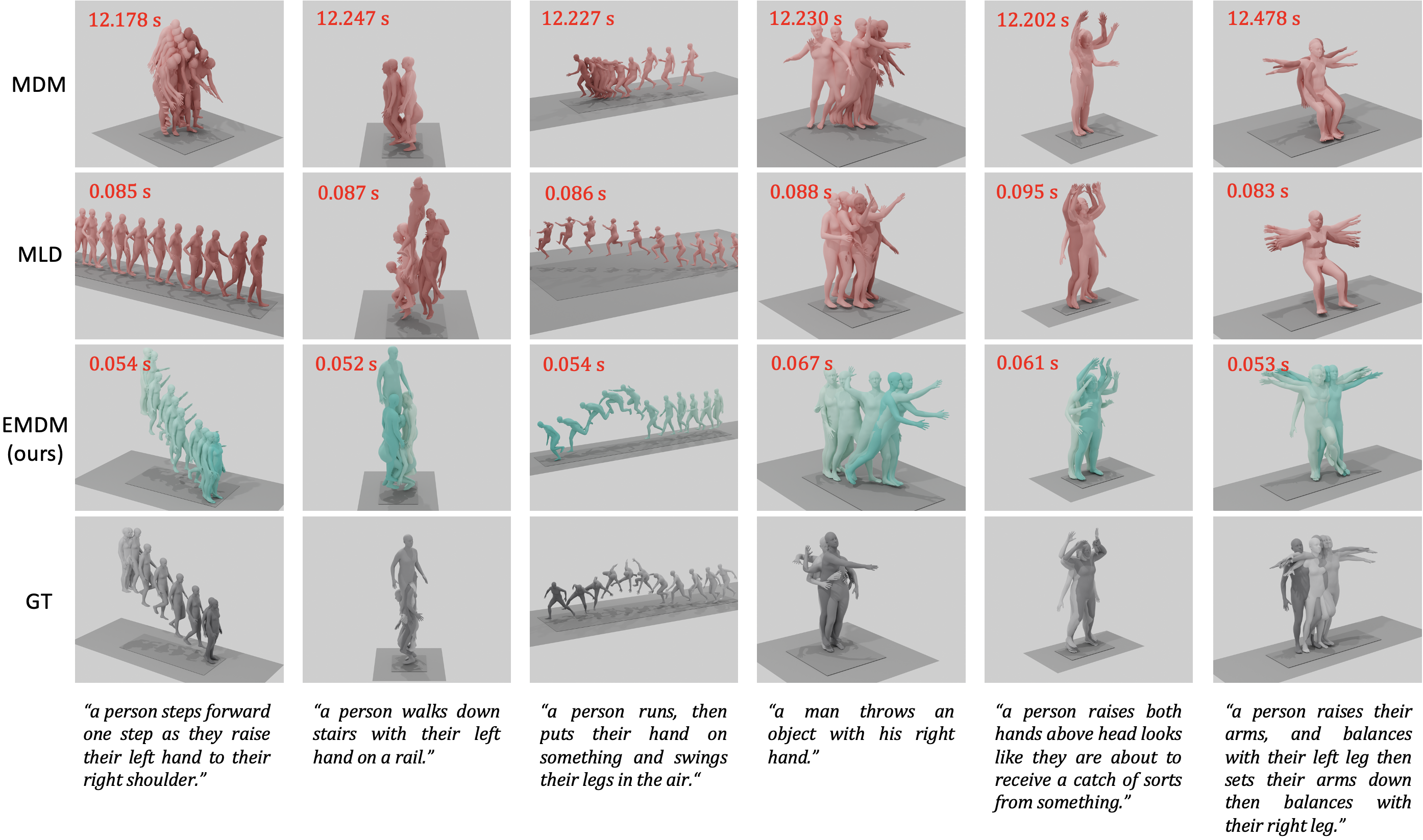}
\caption{Qualitative comparison on text-to-motion task. We visualize the generated motions and real references from six text prompts. 
EMDM achieves the fastest motion generation while delivering high-quality movements that align with the text input.}\label{fig:t2m}
\end{figure}

\subsection{Comparisons on Text-to-motion}
\label{sec:comp:text}

\begin{table}[t]
\footnotesize
\centering
\caption{Comparison of text-to-motion task on HumanML3D~\cite{Guo_2022_CVPR_humanml3d}. The right arrow $\rightarrow$ means the closer to real motion, the better.}
\label{tab:tm:comp:humanml3d}
\resizebox{\textwidth}{!}{
\begin{threeparttable}[b]
\setlength{\tabcolsep}{0.15mm}{
\begin{tabular}{@{}lcccccccc@{}}
\toprule
\multirow{2}{*}{Methods} & \multicolumn{3}{c}{R Precision $\uparrow$}              & \multicolumn{1}{c}{\multirow{2}{*}{FID$\downarrow$}} & \multirow{2}{*}{MM Dist$\downarrow$}              & \multirow{2}{*}{Diversity$\rightarrow$}           
& \multirow{2}{*}{MModality$\uparrow$} 
& \multirow{2}{*}{\makecell[c]{Running Time\\(per frame; ms)$\downarrow$}} 
\\ \cmidrule(lr){2-4}
              & \multicolumn{1}{c}{Top 1} & \multicolumn{1}{c}{Top 2} & \multicolumn{1}{c}{Top 3} & \multicolumn{1}{c}{}                     &                          &                            &                            \\ \midrule
Real &
  $0.511^{\pm.003}$ &
  $0.703^{\pm.003}$ &
  $0.797^{\pm.002}$ &
  $0.002^{\pm.000}$ &
  $2.974^{\pm.008}$ &
  $9.503^{\pm.065}$ &
  \multicolumn{1}{c}{-} &
  -
  \\ \midrule

TEMOS~\cite{petrovich22temos}&
  $0.424^{\pm.002}$ &
  $0.612^{\pm.002}$ &
  $0.722^{\pm.002}$ &
  $3.734^{\pm.028}$ &
  $3.703^{\pm.008}$ &
  $8.973^{\pm.071}$ &
  $0.368^{\pm.018}$  &
  -\\

T2M~\cite{Guo_2022_CVPR_humanml3d}&
  $0.457^{\pm.002}$ &
  $0.639^{\pm.003}$ &
  $0.740^{\pm.003}$ &
  $1.067^{\pm.002}$ &
  $3.340^{\pm.008}$ &
  $9.188^{\pm.002}$ &
  $2.090^{\pm.083}$  &
  -\\
MotionDiffuse~\cite{zhang2022motiondiffuse} &
  ${0.491}^{\pm.001}$ &
  ${0.681}^{\pm.001}$ &
   {{${0.782}^{\pm.001}$}} &
  $0.630^{\pm.001}$ &
  {{${3.113}^{\pm.001}$}} &
  \textcolor{top2}{\bm{${9.410}^{\pm.049}$}} &
  $1.553^{\pm.042}$  &
  $38.235^{\pm2.495}$
  \\
MDM~\cite{mdm2022human}&
    $0.418^{\pm.005}$ &
    $0.605^{\pm.005}$ &
    $0.708^{\pm.005}$ &
    $0.508^{\pm.034}$ &
    $3.630^{\pm.023}$ &
    {{$9.373^{\pm.094}$}} &
    \textcolor{top2}{{${2.880}^{\pm.088}$}} &
    $62.505^{\pm.071}$
  \\
MLD~\cite{chen2023executing}$\bm{\dag}$ &
  ${0.481}^{\pm.003}$ &
  ${0.673}^{\pm.003}$ &
  {${0.772}^{\pm.002}$} &
  $0.473^{\pm.013}$ &
  ${3.196}^{\pm.010}$ &
  $9.724^{\pm.082}$ &
  ${2.413}^{\pm.079}$ &
    \textcolor{top2}{\bm{$0.598^{\pm.004}$}}\\

T2M-GPT~\cite{zhang2023t2m}$\bm{\dag}$ &
  ${0.492}^{\pm.003}$ &
  ${0.679}^{\pm.002}$ &
  ${0.775}^{\pm.002}$ &
  {{$0.141^{\pm.005}$}} &
  ${3.121}^{\pm.009}$ &
  $9.722^{\pm.082}$ &
    ${1.831}^{\pm.048}$ &
{{$0.886^{\pm.007}$}}\\

MoFusion~\cite{dabral2023mofusion} &
 $0.492^{\pm.000}$&
  $-$ &
  $-$ &
  $-$ &
  $-$ &
  $8.820^{\pm.000}$ &
  {{$2.521^{\pm.000}$}} &
{Not open source}\\
    
M2DM~\cite{kong2023priority}$\bm{\dag}$ &
   {{$0.497^{\pm.003}$}} &
   {{$0.682^{\pm.002}$}} &
    $0.763^{\pm.003}$&
  $0.352^{\pm.005}$&
  $3.134^{\pm.010}$ &
  $9.926^{\pm.073}$ &
   \textcolor{top1}{\bm{$3.587^{\pm.072}$}} &
{Not open source}\\

ReMoDiffuse~\cite{zhang2023remodiffuse}$\bm{\dag \ddag}$ &
  \textcolor{top1}{\bm{$0.510^{\pm.005}$}} &
  \textcolor{top1}{\bm{$0.698^{\pm.006}$}} &
  \textcolor{top1}{\bm{$0.795^{\pm.004}$}} &
  \textcolor{top1}{\bm{$0.103^{\pm.004}$}} &
  \textcolor{top1}{\bm{$2.974^{\pm.016}$}} &
  $9.018^{\pm.075}$ &
  $1.795^{\pm.043}$ &
  $0.959^{\pm.002}$ 
    \\

   \midrule
EMDM (Ours) &  
   \textcolor{top2}{\bm{$0.498^{\pm.007}$}} & \textcolor{top2}{\bm{$0.684^{\pm.006}$}} & \textcolor{top2}{\bm{$0.786^{\pm.006}$}} & \textcolor{top2}{\bm{$0.112^{\pm.019}$}} & \textcolor{top2}{\bm{$3.110^{\pm.027}$}} & \textcolor{top1}{\bm{$9.551^{\pm.078}$}} & $1.641^{\pm.078}$ & \textcolor{top1}{\bm{$0.280^{\pm.002}$}} \\

   \bottomrule
\end{tabular}%
}
\begin{tablenotes}
     \item\textcolor{top1}{\textbf{$\text{Blue}$}} and \textcolor{top2}{\textbf{$\text{orange}$}}  indicate the best and the second best result.\\
     $\dag$~\textbf{Two-stage and non end-to-end approach.}\\
    $\ddag$~\textbf{Reference dataset required at the inference stage.}
\end{tablenotes}
\end{threeparttable}
}
\end{table}

\begin{table}[t]
\footnotesize
\caption{Comparison of text-conditional motion generation on KIT~\cite{Plappert2016kit}.}
\label{tab:tm:comp:kit}
\centering
\centering
\resizebox{\textwidth}{!}{
\begin{threeparttable}
\setlength{\tabcolsep}{0.15mm}{
\begin{tabular}{@{}lcccccccc@{}}
\toprule
\multirow{2}{*}{Methods} & \multicolumn{3}{c}{R Precision $\uparrow$}                                                                                                                & \multicolumn{1}{c}{\multirow{2}{*}{FID$\downarrow$}} & \multirow{2}{*}{MM Dist$\downarrow$}              & \multirow{2}{*}{Diversity$\rightarrow$}           & \multirow{2}{*}{MModality$\uparrow$}    & \multirow{2}{*}{\makecell[c]{Running Time\\(per frame; ms)$\downarrow$}} 

\\ \cmidrule(lr){2-4}
              & \multicolumn{1}{c}{Top 1} & \multicolumn{1}{c}{Top 2} & \multicolumn{1}{c}{Top 3} & \multicolumn{1}{c}{}                     &                          &                            &                            \\ \midrule
Real &
  $0.424^{\pm.005}$ &
  $0.649^{\pm.006}$ &
  $0.779^{\pm.006}$ &
  $0.031^{\pm.004}$ &
  $2.788^{\pm.012}$ &
  $11.08^{\pm.097}$ &
  \multicolumn{1}{c}{-}&
- 
  \\ \midrule
TEMOS&
    $0.353^{\pm.006}$ & 
    $0.561^{\pm.007}$ & 
    $0.687^{\pm.005}$ & 
    $3.717^{\pm.051}$ & 
    $3.417^{\pm.019}$ & 
    $10.84^{\pm.100}$ & 
    $0.532^{\pm.034}$ &
- \\
T2M&
  $0.370^{\pm.005}$ &
  $0.569^{\pm.007}$ &
  $0.693^{\pm.007}$ &
  $2.770^{\pm.109}$ &
  $3.401^{\pm.008}$ &
  ${10.91}^{\pm.119}$ &
  $1.482^{\pm.065}$ &
-  \\
MotionDiffuse &
{{${0.417}^{\pm.004}$}}&
${0.621}^{\pm.004}$ &
${0.739}^{\pm.004}$ &
$1.954^{\pm.062}$ &
{{${2.958}^{\pm.005}$}}&
\textcolor{top1}{\bm{${11.10}^{\pm.143}$}} &
$0.730^{\pm.013}$&
$68.403^{\pm6.982}$  \\
MDM&
  $0.405^{\pm.007}$ &
$0.610^{\pm.007}$ &
$0.732^{\pm.007}$ &
$0.508^{\pm.030}$ &
$3.085^{\pm.022}$&
$10.74^{\pm.096}$ &
{{${1.834}^{\pm.052}$}} &
${64.636}^{\pm.463}$ \\

MLD~\cite{chen2023executing}$\bm{\dag}$ &
${0.390}^{\pm.008}$ & 
${0.609}^{\pm.008}$ & 
${0.734}^{\pm.007}$ & 
{{${0.404}^{\pm.027}$}} & 
${3.204}^{\pm.027}$ & 
$10.80^{\pm.117}$ & 
\textcolor{top2}{\bm{${2.192}^{\pm.071}$}} &
    \textcolor{top2}{\bm{$0.673^{\pm.008}$}}
 \\
T2M-GPT~\cite{zhang2023t2m}$\bm{\dag}$ &
  ${0.416}^{\pm.006}$ &
  ${0.627}^{\pm.006}$ &
{{${0.745}^{\pm.006}$}} &
  $0.514^{\pm.029}$ &
  ${3.007}^{\pm.023}$ &
{{$10.92^{\pm.108}$}} &
  ${1.570}^{\pm.039}$ &
{{$0.905^{\pm.011}$}}\\
M2DM~\cite{kong2023priority}$\bm{\dag}$ &
  $0.416^{\pm.004}$ &
{{$0.628^{\pm.004}$}} &
  $0.743^{\pm.004}$ &
  $0.515^{\pm.029}$ &
  $3.015^{\pm.017}$ &
  $11.42^{\pm.970}$ &
  \textcolor{top1}{\bm{$3.325^{\pm.37}$}} &
  {Not open source}\\

ReMoDiffuse~\cite{zhang2023remodiffuse}$\bm{\dag\ddag}$ &
  \textcolor{top2}{\bm{$0.427^{\pm.014}$}} &
   \textcolor{top2}{\bm{$0.641^{\pm.004}$}} &
   \textcolor{top2}{\bm{$0.765^{\pm.055}$}} &
  \textcolor{top1}{\bm{$0.155^{\pm.006}$}} &
  \textcolor{top1}{\bm{$2.814^{\pm.012}$}} &
  $10.80^{\pm.105}$ &
  $1.239^{\pm.028}$ &
  $1.002^{\pm.007}$ 
    \\
 
\midrule

EMDM (Ours) &
    \textcolor{top1}{\bm{$0.443^{\pm.006}$}} & \textcolor{top1}{\bm{$0.660^{\pm.006}$}} & \textcolor{top1}{\bm{$0.780^{\pm.005}$}} & \textcolor{top2}{\bm{$0.261^{\pm.014}$}} & \textcolor{top2}{\bm{$2.874^{\pm.015}$}} & \textcolor{top2}{\bm{$10.96^{\pm.093}$}} & 
    $1.343^{\pm.089}$ & \textcolor{top1}{\bm{$0.284^{\pm.002}$}}
   \\
   \bottomrule
\end{tabular}
}
\end{threeparttable}
}
\end{table}

We evaluate EMDM on the text-to-motion task. We use the frozen CLIP~\cite{radford2021learning} model as $\tau_\theta^{w}$ to encode the text, giving $w_{clip}^{1}\in \mathbb{R}^{1,024}$. The motion is then synthesized by conditioning on text input $\rvc = \{w^{1:N}\}$.  We compare our model with SOTA methods on HumanML3D and KIT with the metrics proposed by~\cite{Guo_2022_CVPR_humanml3d}. \cref{tab:tm:comp:humanml3d} and \cref{tab:tm:comp:kit} summarize the comparison results on HumanML3D~\cite{Guo_2022_CVPR_humanml3d} and KIT dataset~\cite{Plappert2016kit}, respectively. 
In Tab.~\ref{tab:tm:comp:humanml3d}, EMDM demonstrates the highest motion generation speed and highly competitive performance compared with the very recent reference-based ~\cite{zhang2023remodiffuse} approach across all metrics, which validates its effectiveness and efficiency. In Tab.~\ref{tab:tm:comp:kit}, EMDM consistently outperforms all existing methods in motion generation speed and Top1/2/3 matching accuracy. It produces competitive results on the remaining metrics. To summarize, EMDM demonstrates significant advantages compared to other methods, including multi-stage ones. Note that EMDM is single-stage and end-to-end trainable. Although ReMoDiffuse~\cite{zhang2023remodiffuse} 
attains competitive speeds in motion generation with relatively high quality, it employs a retrieval-based approach that relies on reference motion datasets for generating motions at the inference stage. ReMoDiffuse is also a two-stage method that is non-end-to-end trainable. We provide qualitative results in Fig.~\ref{fig:t2m}, where EMDM achieves the fastest motion generation while maintaining competitive motion quality.

\subsection{Comparisons on Action-to-motion}
\label{sec:comp:action}

\begin{wrapfigure}{r}{6cm}
\vspace{-12mm}
  \begin{center}
\includegraphics[width=\linewidth]{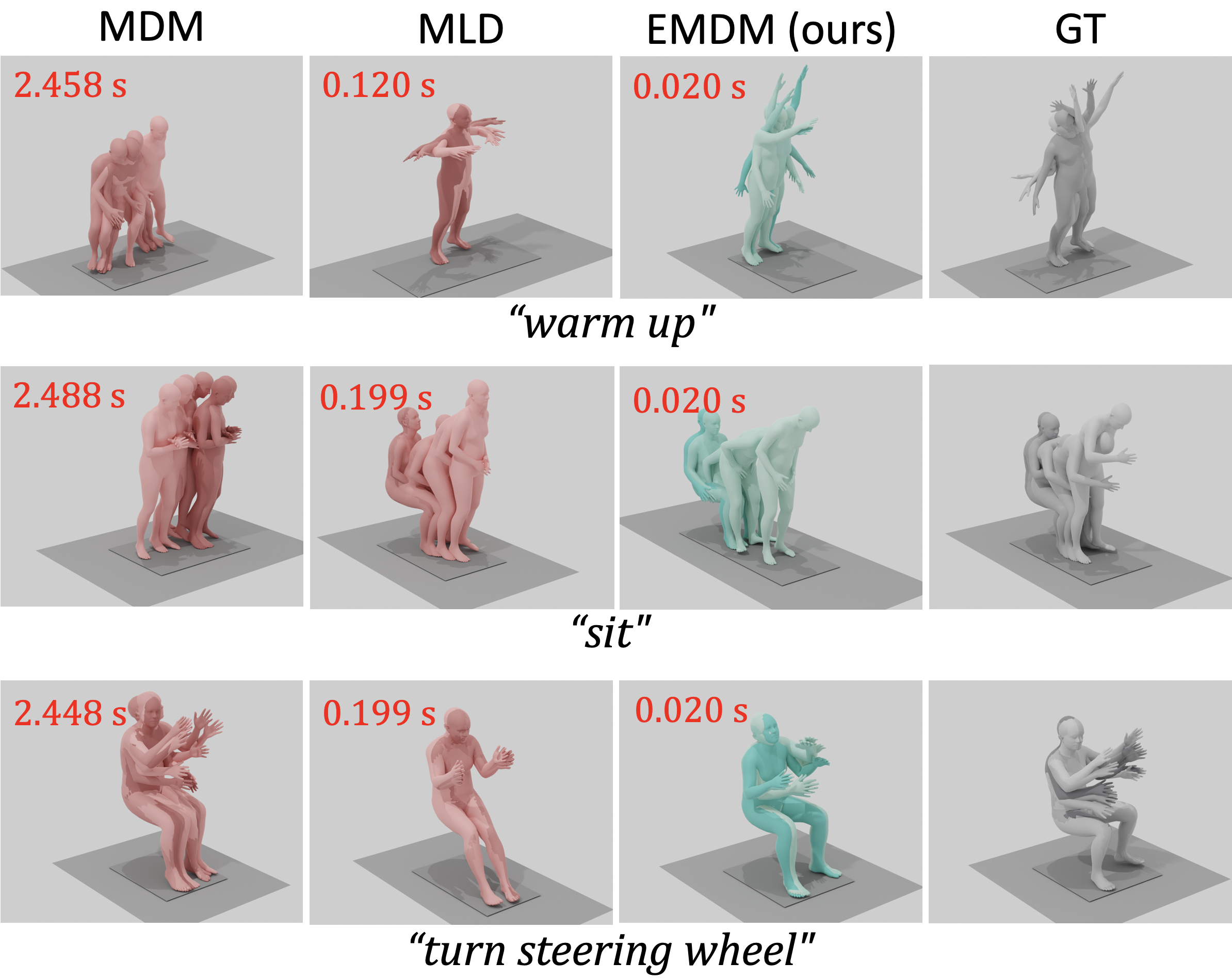}
  \end{center}
  \vspace{-5mm}
\caption{Qualitative comparisons on action-to-motion task.}
  \label{fig:a2m}
  \vspace{-10mm}
\end{wrapfigure}

The action-conditioned task is to generate human motion given an action label. Following~\cite{mdm2022human,chen2023executing}, we report the FID, ACC, DIV, MM and Running Time of the aforementioned methods. The comparison on HumanAct12~\cite{guo2020action2motion} is shown in \cref{tab:comp:action}. EMDM achieves competitive results on HumanAct12 while achieving superior run-time performance. Notably, although MLD also uses less time for motion sampling, it is a two-stage method. The qualitative comparison of action-to-motion is visualized in Fig.~\ref{fig:a2m}, where EMDM achieves efficient motion generation performance while aligning with the semantics of the action label, while others have improper motion semantics, such as the ``sit'' motion of MDM and stiff "turn steering wheel" motion of MLD. See More results in the supplementary video.
\begin{table}[t]
\centering
\caption{Comparison of action-to-motion task on HumanAct12~\cite{guo2020action2motion}: $\text{FID}_\text{train}$ indicating the evaluated splits. Accuracy (ACC) for action recognition. Diversity (DIV) and MModality (MM) for generated motion diversity w.r.t each action label.}
\resizebox{\textwidth}{!}{
\begin{threeparttable}[b]
\setlength{\tabcolsep}{4mm}{
\begin{tabular}{lccccc}
\toprule
{Methods} &  \multicolumn{1}{c}{$\text{FID}_{\text{train}}\downarrow$} & ACC $\uparrow$& DIV$\rightarrow$ & MM$\rightarrow$ & \makecell[c]{Running Time\\(per frame; ms)$\downarrow$} \\ 
\toprule                
Real &
  $0.020^{\pm.010}$ &
  $0.997^{\pm.001}$ &
  $6.850^{\pm.050}$ &
  $2.450^{\pm.040}$ & -
  \\
  \midrule
ACTOR \cite{petrovich21actor}&
  $0.120^{\pm.000}$ &
  $0.955^{\pm.008}$ &
    \textcolor{top1}{\bm{$6.840^{\pm.030}$}} &
  $2.530^{\pm.020}$  & 
  $0.523^{\pm.009}$
  
  \\
INR \cite{cervantes2022implicit}&
  $0.088^{\pm.004}$ &
  $0.973^{\pm.001}$ &
$6.881^{\pm.048}$&
  $2.569^{\pm.040}$ & -
  
  \\
MDM \cite{mdm2022human}&
  $0.100^{\pm.000}$ &
  \textcolor{top2}{\bm{$0.990^{\pm.000}$}}&
  \textcolor{top1}{\bm{$6.860^{\pm.050}$}} &
  \textcolor{top2}{\bm{${2.520}^{\pm.010}$}} & 
  $41.154^{\pm .162}$
  \\
MLD~\cite{chen2023executing}$\bm{^\dag}$ &
\textcolor{top1}{\bm{${0.077}^{\pm0.004}$}}&
$0.964^{\pm.002}$&
${6.831}^{\pm0.050}$&
$2.824^{\pm.038}$ &
\textcolor{top2}{\bm{$1.998^{\pm.001}$}}

\\
      \midrule
EMDM (Ours) &
\textcolor{top2}{\bm{${0.084^{\pm.004}}$}} & 
\textcolor{top1}{\bm{$0.991^{\pm.003}$}}	& 
\textcolor{top2}{\bm{$6.876^{\pm.148}$}}&
\textcolor{top1}{\bm{$2.417^{\pm 1.009}$}} & 
\textcolor{top1}{\bm{$0.337^{\pm .005}$ }} 
\\ \bottomrule
\end{tabular}
}
\begin{tablenotes}
     \item\textcolor{top1}{\textbf{$\text{Blue}$}} and \textcolor{top2}{\textbf{$\text{orange}$}} indicate the best and the second best result. \\
     $\dag$~\textbf{Two-stage and non end-to-end approach.}
\end{tablenotes}
\end{threeparttable}
}
\label{tab:comp:action}
\end{table}

\section{Ablation Study}
\label{sec:ablation}

We validate the effectiveness of our key design choices in the following, with all experiments tested on HumanML3D~\cite{Guo_2022_CVPR_humanml3d} as text-to-motion is a more challenging task, compared to action-to-motion. The number of frames of the generated motion is $196$. All models are trained with the same training settings. We also study the performance of the model when trained without providing conditions to the discriminator with geometric loss in Appendix~\redlink{D3}.

\begin{figure}[t]
    \centering
    \includegraphics[width=\linewidth]{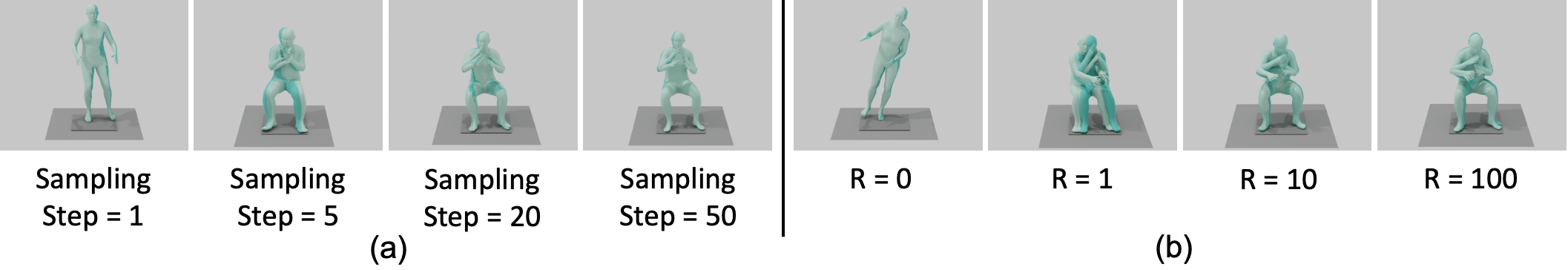}
    \caption{Ablation studies on different sample steps~(a) and weights of geometric loss~(b) of generated motions. We use a classic textual description, "sit", as the input condition.}
    \label{fig:ablation_study}
\end{figure}

\subsection{Influence of the Number of Sampling Steps}
\label{sec:ablation_step_size}
We investigate the influence of different sampling steps on the performance. We train and test our model with sampling step numbers $1, 5, 10, 20$ and $50$. Notably, when the step number is set to $1$, the whole model can be regarded as a GAN model. As shown in Tab.~\ref{tab:ablation_sampling_steps_hml}, when increasing the step size, the sampling speed is improved significantly. However, when the step size is too large, the motion quality indicated by FID, DIV, and MM drops. This is also witnessed by the qualitative results in Fig.~\ref{fig:ablation_study}~(a), where increasing sampling steps
promote motion semantics, i.e., "sit". We consistently set the sampling step size to $10$ in the experiments. 

\begin{table}[t]
        \centering
\caption{Influence of sampling steps on motion generation using HumanML3D.}
\label{tab:ablation_sampling_steps_hml}
\resizebox{\textwidth}{!}{
\setlength{\tabcolsep}{0.15mm}{
\begin{tabular}{@{}lcccccccc@{}}
\toprule
\multirow{2}{*}{\makecell[c]{\#Steps}} & \multicolumn{3}{c}{R Precision $\uparrow$}              & \multicolumn{1}{c}{\multirow{2}{*}{FID$\downarrow$}} & \multirow{2}{*}{MM Dist$\downarrow$}              & \multirow{2}{*}{Diversity$\rightarrow$}           
& \multirow{2}{*}{MModality$\uparrow$} 
& \multirow{2}{*}{\makecell[c]{Running Time\\(per frame; ms)$\downarrow$}} 
\\ \cmidrule(lr){2-4}
              & \multicolumn{1}{c}{Top 1} & \multicolumn{1}{c}{Top 2} & \multicolumn{1}{c}{Top 3} & \multicolumn{1}{c}{}                     &                          &                            &                            \\ \midrule
Real &
  $0.511^{\pm.003}$ &
  $0.703^{\pm.003}$ &
  $0.797^{\pm.002}$ &
  $0.002^{\pm.000}$ &
  $2.974^{\pm.008}$ &
  $9.503^{\pm.065}$ &
  \multicolumn{1}{c}{-} &
  -
  \\ \midrule

$1$ & $0.345^{\pm.005}$ & $0.525^{\pm.007}$ & $0.645^{\pm.007}$ & $5.640^{\pm.127}$ & $4.278^{\pm.021}$ & $7.639^{\pm.071}$ & ${0.622^{\pm.016}}$ & \textcolor{top1}{\bm{$0.004^{\pm.000}$}} \\

$5$ & $0.368^{\pm.005}$ & $0.547^{\pm.006}$ & $0.655^{\pm.006}$ & $1.306^{\pm.052}$ & $4.047^{\pm.025}$ & $9.168^{\pm.074}$ & \textcolor{top1}{\bm{$2.285^{\pm.065}$}} & $0.152^{\pm.000}$ \\

$\textbf{10}$ &  \textcolor{top1}{\bm{$0.498^{\pm.007}$}} & \textcolor{top1}{\bm{$0.684^{\pm.006}$}} & \textcolor{top1}{\bm{$0.786^{\pm.006}$}} & \textcolor{top1}{\bm{$0.112^{\pm.019}$}} & \textcolor{top1}{\bm{$3.110^{\pm.027}$}} & $9.551^{\pm.078}$ & $1.641^{\pm.078}$ & $0.280^{\pm.002}$ \\

$20$ & $0.490^{\pm.006}$ & $0.679^{\pm.005}$ & $0.780^{\pm.005}$ & $0.191^{\pm.028}$ & $3.142^{\pm.023}$ & $9.531^{\pm.074}$ & $1.688^{\pm.057}$ & $0.555^{\pm.002}$ \\

$50$ & $0.479^{\pm.007}$ & $0.671^{\pm.007}$ & $0.770^{\pm.005}$ & $0.216^{\pm.027}$ & $3.168^{\pm.028}$ & \textcolor{top1}{\bm{$9.482^{\pm.083}$}} & $1.788^{\pm.046}$ & $1.356^{\pm.000}$ \\

   \bottomrule
\end{tabular}
}
}
\end{table}

\subsection{Influence of Geometric Loss}
\label{sec:ablation_geometric_loss}
We study the influence of geometric loss during EMDM training.
Recall the overall loss for our condition generator is denoted as $\mathcal{L} = \mathcal{L}_{\text{disc}} + R\cdot \mathcal{L}_{\text{geo}}$ (Eq.~\ref{eq:total_loss}), where $\mathcal{L}_{\text{disc}}$ and $\mathcal{L}_{\text{geo}}$ represent the generator loss and geometric losses, respectively. Here, $R$ serves as a balancing term. We evaluate the motion quality and running time by setting $R$ to be $0.0, 1.0,10.0,100.0$ in Eq.~\ref{eq:total_loss}. As shown in Tab.~\ref{tab:ablation_weights_humanml3d}, when no geometric loss is applied, the motion quality significantly drops, e.g., FID $= 9.308$. Meanwhile, imposing geometric loss effectively improves the motion quality during the training process. We visualize the human motion under different weights $R$ in Fig.~\ref{fig:ablation_study} (b).
In this paper, we empirically set the $R$ to be $100.0$ for text-to-motion tasks and $1.0$ for action-to-motion tasks. 

\begin{table}[t]
      \centering
\caption{Influence of geometric loss weights on motion generation using HumanML3D.}
\label{tab:ablation_weights_humanml3d}
\resizebox{\textwidth}{!}{
\setlength{\tabcolsep}{2mm}{
\begin{tabular}{@{}lccccccc@{}}
\toprule
\multirow{2}{*}{\makecell[c]{R value}} & \multicolumn{3}{c}{R Precision $\uparrow$}              & \multicolumn{1}{c}{\multirow{2}{*}{FID$\downarrow$}} & \multirow{2}{*}{MM Dist$\downarrow$}              & \multirow{2}{*}{Diversity$\rightarrow$}           
& \multirow{2}{*}{MModality$\uparrow$} 

\\ \cmidrule(lr){2-4}
              & \multicolumn{1}{c}{Top 1} & \multicolumn{1}{c}{Top 2} & \multicolumn{1}{c}{Top 3} & \multicolumn{1}{c}{}                     &                          &                            &                            \\ \midrule
Real &
  $0.511^{\pm.003}$ &
  $0.703^{\pm.003}$ &
  $0.797^{\pm.002}$ &
  $0.002^{\pm.000}$ &
  $2.974^{\pm.008}$ &
  $9.503^{\pm.065}$ &
  \multicolumn{1}{c}{-} 
  \\ \midrule

$0$ & $0.197^{\pm.005}$ & $0.338^{\pm.006}$ & $0.445^{\pm.006}$ & $9.308^{\pm.190}$ & $5.463^{\pm.027}$ & $8.337^{\pm.086}$ & \textcolor{top1}{\bm{$3.140^{\pm.079}$}} \\

$1$ & $0.468^{\pm.006}$ & $0.656^{\pm.004}$ & $0.761^{\pm.003}$ & $0.449^{\pm.047}$ & $3.272^{\pm.018}$ & $9.445^{\pm.084}$ & $1.978^{\pm.065}$ \\

$10$ & $0.486^{\pm.005}$ & $0.672^{\pm.004}$ & $0.768^{\pm.005}$ & $0.232^{\pm.034}$ & $3.169^{\pm.025}$ & $9.347^{\pm.076}$ & $1.706^{\pm.037}$ \\

\bm{$100$} &  \textcolor{top1}{\bm{$0.498^{\pm.007}$}} & $0.684^{\pm.006}$ & \textcolor{top1}{\bm{$0.786^{\pm.006}$}} & \textcolor{top1}{\bm{$0.112^{\pm.019}$}} & \textcolor{top1}{\bm{$3.110^{\pm.027}$}} & \textcolor{top1}{\bm{$9.551^{\pm.078}$}} & $1.641^{\pm.078}$  \\

$1000$ & $0.494^{\pm.005}$ & \textcolor{top1}{\bm{$0.685^{\pm.004}$}} & {$0.778^{\pm.005}$} & $0.195^{\pm.026}$ & {$3.120^{\pm.022}$} & $9.595^{\pm.084}$ & $1.600^{\pm.045}$ \\

   \bottomrule
\end{tabular}
}
}
\end{table}

\section{Conclusion}

In this paper, we reveal efficiency issues with the existing motion diffusion models and the challenges in accelerating the models. We introduce the Efficient Motion Diffusion Model (EMDM) to overcome the obstacles faced by existing generative diffusion models in achieving fast and high-quality motion generation. Different from previous approaches, we propose to sample motion from a diffusion model with much fewer sampling steps at the denoising stage. We utilize a conditional denoising diffusion Generative Adversarial Network to model the complex denoising distributions conditioning on the control signals. This enables the use of much larger step sizes, which in turn reduces the number of sampling steps while maintaining high motion quality and consistency in semantics with respect to the condition. We also incorporate a geometric loss to further elevate motion quality and enhance training efficiency. The whole model is end-to-end trainable. Consequently, EMDM achieves a remarkable speed-up without sacrificing motion quality when compared to current motion diffusion models, demonstrating its efficiency and effectiveness.\\
\textit{Limitations and Future Works.}
Although EMDM demonstrates encouraging performance in efficient human motion generation, its motion generation process lacks physical considerations, which may lead to issues like floating and ground penetration; See Fig.~\redlink{E5} in Appendix~\redlink{E}. Efforts to integrate physics-based characters\cite{yuan2023physdiff, jiang2023drop, dou2023c, 2021-TOG-AMP, peng2022ase, peng2018deepmimic} show promise for future improvements. In addition, currently EMDM accepts mainly textual inputs, but its potential extends to visual inputs~\cite{li2021hybrik, rempe2021humor, liao2024vinecs, chen2022learning, zhang2023learning, dou2023tore, wang2023zolly,shi2020motionet,sun2024aios} or music sources~\cite{alexanderson2023listen,li2022danceformer, lee2019dancing} for online motion synthesis, offering other exciting research directions.
\section*{Acknowledgements} 
The authors would like to thank Biao Jiang, Zhengyi Luo, Weilin Wan, and Chen Wang, as well as the AnySyn3D Team and the ECIG Team, for their engaging and insightful discussions. This work is partly supported by the Innovation and Technology Commission of the HKSAR Government under the ITSP-Platform grant (Ref:  ITS/319/21FP) and the InnoHK initiative (TransGP project).

\clearpage
\appendix
\renewcommand\thefigure{\Alph{section}\arabic{figure}}   
\renewcommand\thetable{\Alph{section}\arabic{table}}

This supplementary material covers: More Qualitative Results~(Sec.~\ref{sec:supp_more_results}); Unconditional Motion Generation~(Sec.~\ref{sec:supp_uncond_gen}); Implementation Details~(Sec.~\ref{sec:supp_implementation_details}) and More Experimental Results~(Sec.~\ref{sec:supp_more_exp}). Please watch our supplementary video for a more thorough review.

\section{More Qualitative Results}
\label{sec:supp_more_results}
\begin{figure}
    \centering
\includegraphics[width=0.92\linewidth]{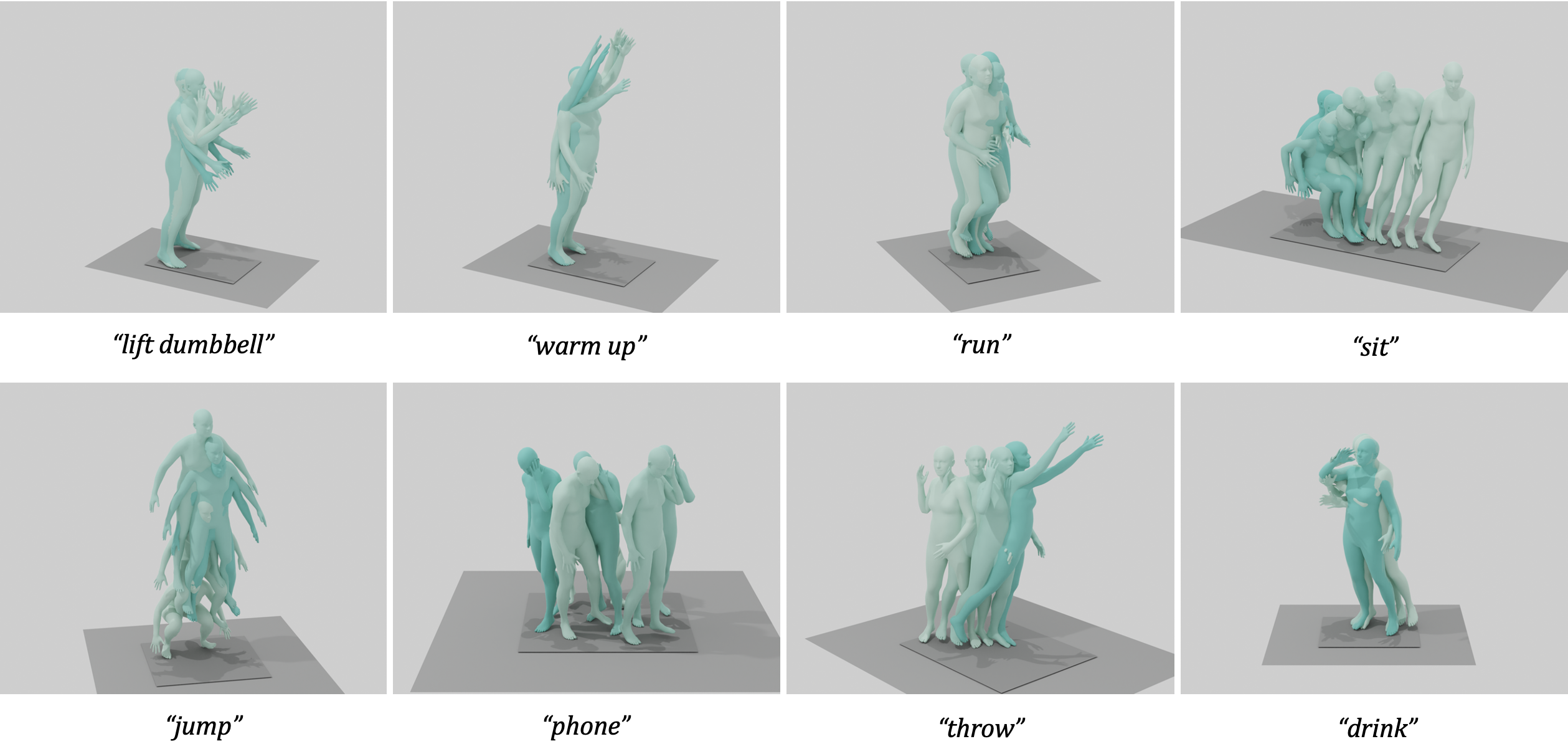}
\caption{More qualitative results of EMDM on the task of action-to-motion.}
    \label{fig_supp:more_a2m}
\end{figure}
\begin{figure}[H]
    \centering
\includegraphics[width=0.92\linewidth]{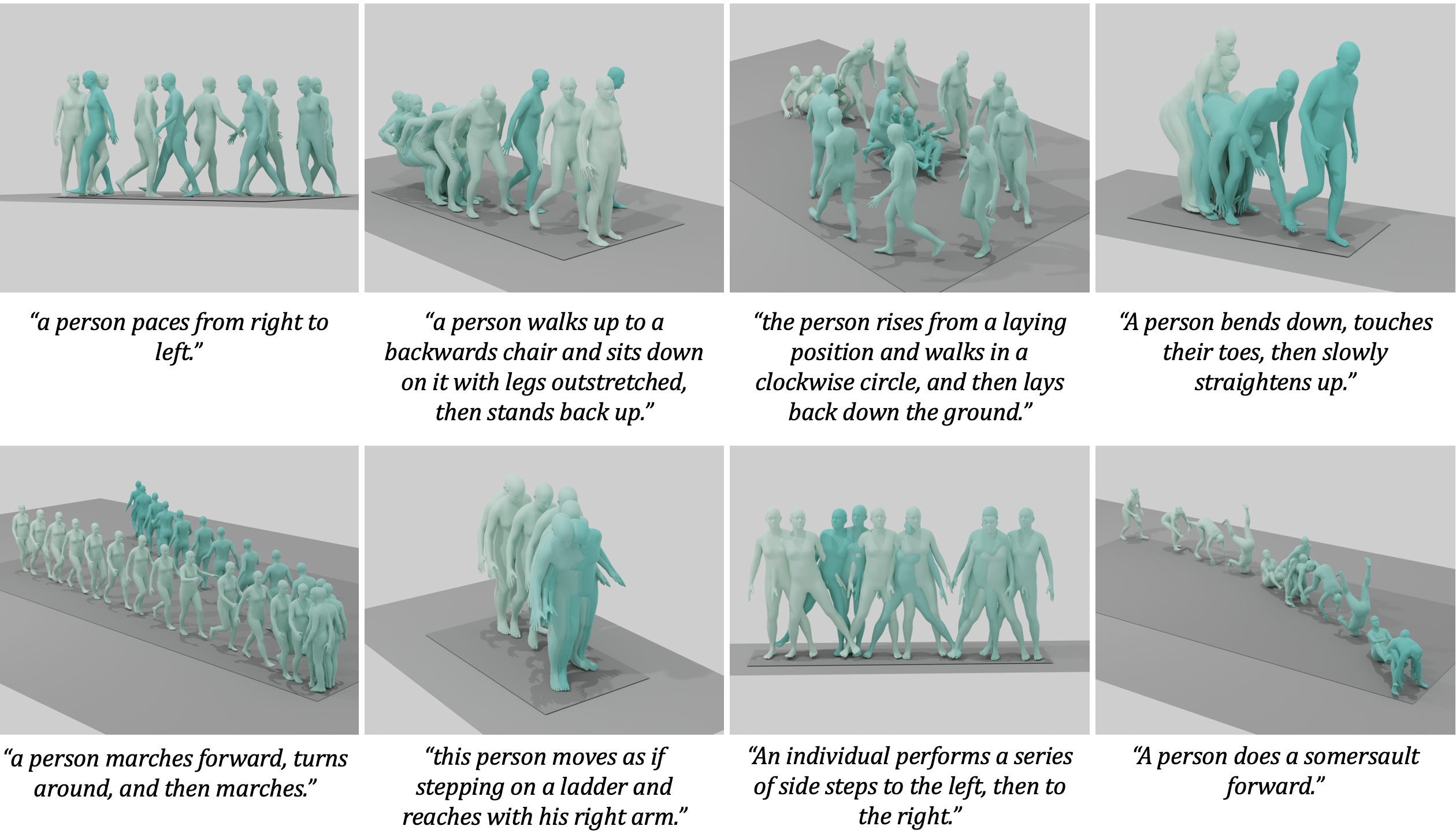}
\caption{More qualitative results of EMDM on the task of text-to-motion.}
    \label{fig_supp:more_t2m}
\end{figure}

In the following, we provide more qualitative results of action-to-motion and text-to-motion tasks, which are visualized in Fig.~\ref{fig_supp:more_a2m} and Fig.~\ref{fig_supp:more_t2m}. The model is evaluated on the HumanAct12~\cite{guo2020action2motion} dataset and HumanML3D dataset~\cite{Guo_2022_CVPR_humanml3d}, respectively. EMDM produces high-quality human motions that faithfully align with the input conditions. We highly suggest readers watch our supplementary video for a more thorough review.

\section{Unconditional Motion Generation}
\label{sec:supp_uncond_gen}
Next, we evaluate unconditional motion generation following~\cite{chen2023executing}. As shown in Tab.~\ref{tab:comp:uncondition}, EMDM exhibits higher motion quality and significantly reduced running time when compared to existing methods.
\begin{table}[H]
\centering
\caption{Comparison of unconditional motion generation task on the part of AMASS dataset following~\cite{chen2023executing}.}
\resizebox{\textwidth}{!}{
\setlength{\tabcolsep}{6mm}{
\begin{threeparttable}
\begin{tabular}{lccc}
\toprule
{Methods} &  \multicolumn{1}{c}{$\text{FID}\downarrow$} & \makecell[c]{Diversity$\rightarrow$} & Running Time~(per frame; ms)$\downarrow$\\ 
\toprule        

Real &
  $0.002$ &
  $9.503$ & -
  \\ \midrule
VPoser-t~\cite{vposer_SMPL-X:2019}&
  $36.65$ & $3.259$ & -
  
  \\

ACTOR~\cite{petrovich21actor}&
  $14.14$ &
  $5.123$ & $0.523^{\pm.009}$
  \\
MDM~\cite{mdm2022human}&
  $8.84$ &
  $6.429$ &
  $62.505^{\pm.071}$
  \\
MLD~\cite{chen2023executing}~$\bm{\dag}$&
  \textcolor{top1}{\bm{$1.4$}} &
  \textcolor{top2}{\bm{$8.577$}} &
  \textcolor{top2}{\bm{$0.886^{\pm.007}$}}
  \\

      \midrule
EMDM (Ours) &
\textcolor{top2}{\bm{$3.46$}} & 
\textcolor{top1}{\bm{$8.759$}}	 &
\textcolor{top1}{\bm{$0.280^{\pm.002}$}}
\\ \bottomrule
\end{tabular}

\begin{tablenotes}
     \item\textcolor{top1}{\textbf{$\text{Blue}$}} and \textcolor{top2}{\textbf{$\text{orange}$}}  indicate the best and the second best result.\\
     $\dag$~\textbf{Two-stage and non end-to-end approach.}\\
\end{tablenotes}
\end{threeparttable}

}
}
\label{tab:comp:uncondition}
\end{table}

\section{Implementation Details}
\label{sec:supp_implementation_details}
In the following, we present the network structures and training details of EMDM. During the training stage, we noise a ground-truth image $\rvx_0$ to $\rvx_{t-1}$ and $\rvx_{t}$ given a time step $t$. We use the $\rvx_{t}$, as well as conditions (text/action $\rvc$, time step $t$) and latent variable $\rvz$ to generate $\hat{\rvx}_0$ which is then used to sample $\hat{\rvx}_{t-1}$. The fake $\hat{\rvx}_{t-1}$ or real $\rvx_{t-1}$, together with conditions (text/action $\rvc$, time step $t$, and the real $\rvx_{t}$), are fed to the conditional discriminator. During inference, conditions (including text/action $\rvc$, time step $t$, and $\rvx_{t}$) and a latent variable $\rvz$ are fed to our generator. The denoised output is the generated motion.

\begin{figure}
\centering
\includegraphics[width=0.8\linewidth]{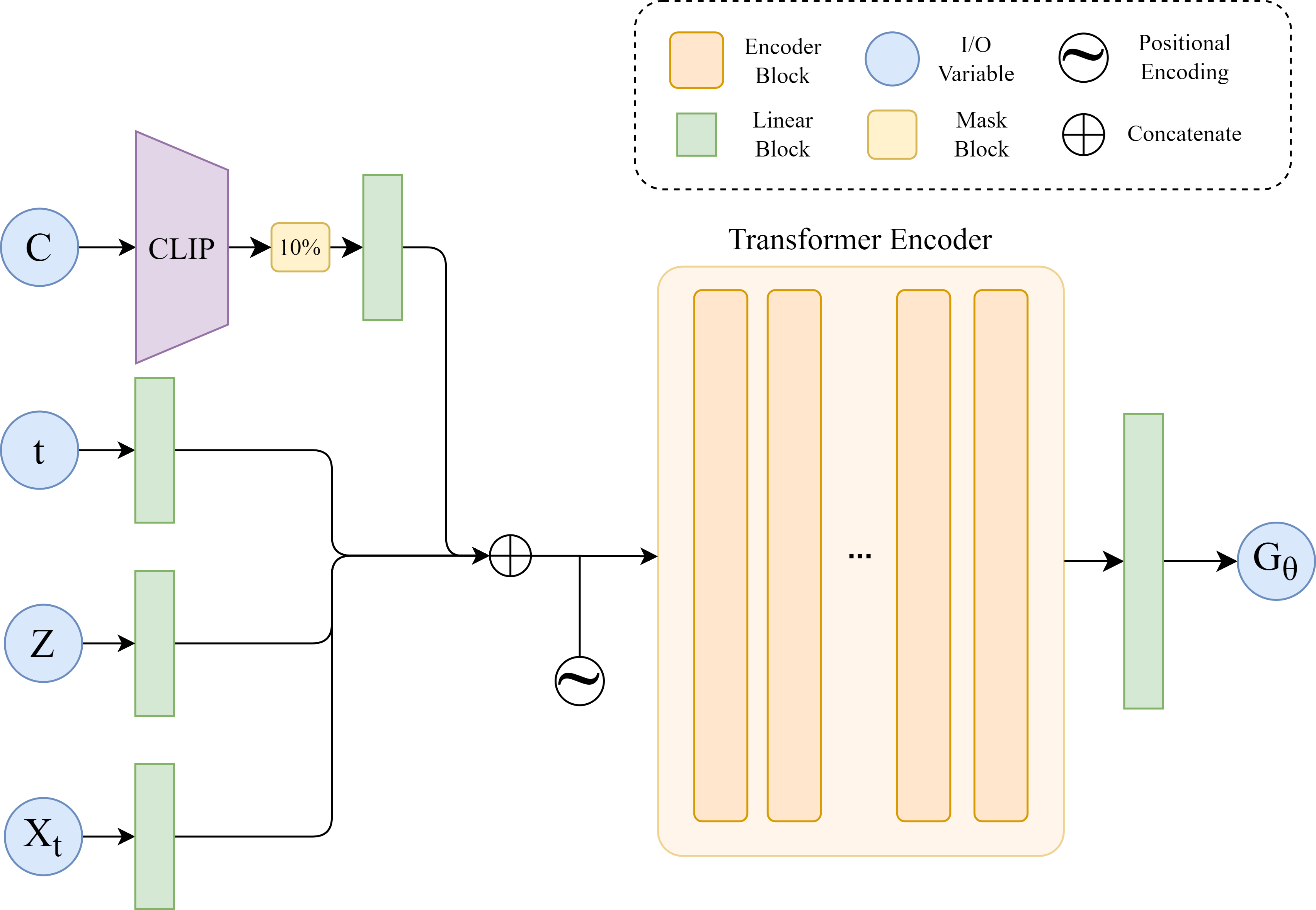}
\caption{The generator architecture for the text-to-motion tasks. For the action-to-motion task, the \textit{CLIP} module, masking module, and the corresponding linear layer are replaced with a single linear layer for action label embedding. The linear layer for $\rvz$ consists of 5 layers.}
    \label{fig_supp:fig_gen}
\end{figure}

\begin{figure}
\centering
\includegraphics[width=0.8\linewidth]
{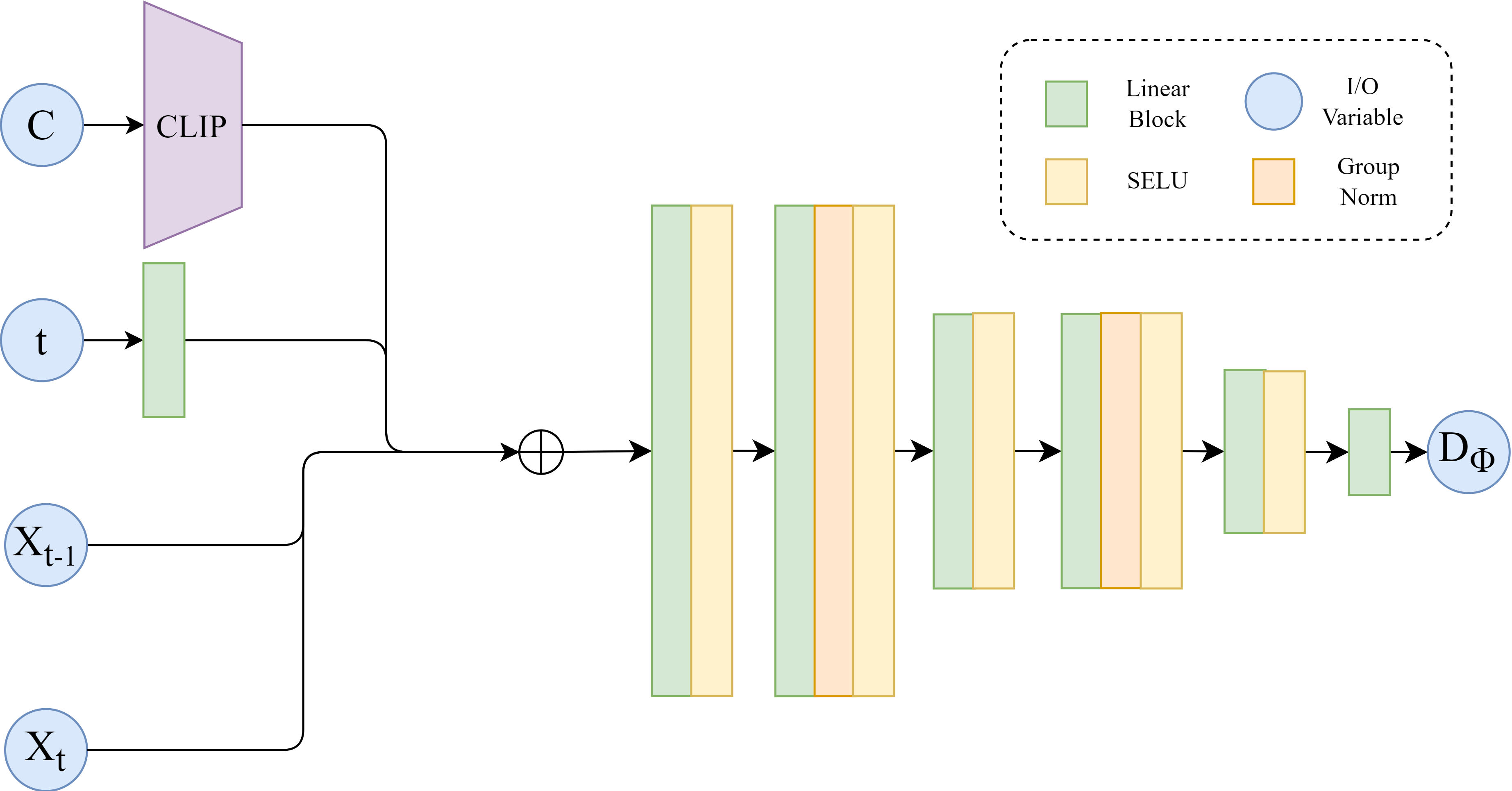}
\caption{The discriminator architecture for the text-to-motion task.  We replace the  \textit{CLIP} module with one-hot encoding for the action-to-motion task.}
\label{fig_supp:fig_disc}
\end{figure}

\subsection{Conditional Generator Structure}
In this paper, we employ a conditional generator for synthesizing motion conditioned on text or action labels, time step $t$ and human motion $\rvx_t$ at $t$-th time step. The model can be written as $G_{\theta}(\rvx_t, \rvz, \rvc, t)$, where $\rvx_t$ is the motion to be denoised, $\rvz \in \mathbb{R}^{64}$ is the latent variable for GAN, and $\rvc$, either a string of text or an action number $\in \mathbb{R}^1$, is the input control signal.  The network structure of $G_{\theta}$ is shown as in Fig.~\ref{fig_supp:fig_gen}.

\paragraph{T2M Architecture}

$t$ and $\rvz$ are mapped to $\mathbb{R}^{1024}$ by $1$ and $5$ linear layers respectively, while $\rvc$ is encoded by \textit{CLIP} \cite{radford2021learning} to $\mathbb{R}^{512}$, randomly masked $10$\% of the values and embedded to $\mathbb{R}^{1024}$ by a linear layer. $\rvx_{t}$ is mapped to $\mathbb{R}^{seq\times1024}$ by a linear layer, where $seq$ is the length of the motion. All the aforementioned values are concatenated and fed to the encoder. We discard the first $3$ tokens of the output and map it back to a motion using a linear layer.

We use the PyTorch implementation for Transformers. The model has $12$ transformer layers and $32$ attention heads. The feed-forward size and latent dimension are both set to be $1024$. The dropout rate is $0.1$. We employ \textit{selu} as the activation function.

\paragraph{A2M Architecture}
The overall architecture is the same. The only difference for action-to-motion tasks is that instead of using \textit{CLIP} + masking + linear layer to map a text to $\mathbb{R}^{1024}$, we use a linear layer to map the action number directly to $\mathbb{R}^{1024}$.

\subsection{Conditional Discriminator Structure}
In EMDM, we employ a conditional discriminator for assessing the authenticity of motions, which can be written as $D_\phi(\rvx_{t-1}, \rvx_t, \rvc, t)$, where $\rvx_{t-1}$ is the motion to be assessed. The input control signals $\rvc$ (either a string of text or an action number $\in \mathbb{R}^1$), time step $t$ and $\rvx_t$ serve as the conditions for $D_\phi$. The network structure is shown in Figure ~\ref{fig_supp:fig_disc}. 

After training, the discriminator would give a positive value for real motions $\rvx_{t-1}$ and a negative value for the fake ones $\hat \rvx_{t-1}$. 

\paragraph{T2M Architecture}
The Discriminator consists of $7$ linear layers, each followed by a \textit{selu} layer. Group normalization is applied after two of the linear layers as well. $t$ is embedded to $\mathbb{R}^{128}$ with sinusoidal positional embeddings as similar to \cite{xiao2021tackling}. $\rvc$ is embedded to $\mathbb{R}^{512}$ using $CLIP$.
We then concatenate ${\rvx}_t$, $\rvx_{t-1}$, embedded $t$ and embedded $\rvc$ and pass the result to the linear layers.

\paragraph{A2M Architecture}
The overall architecture is the same. The only difference for action-to-motion tasks is that instead of using $CLIP$ to embed the text, we use one-hot encoding to transform the action number from $\mathbb{R}^1$ to $\mathbb{R}^{A}$, where $A$ is the number of possible action labels.

\subsection{Training Details}
During network training, we adopt the scheduling scheme following \cite{xiao2021tackling}. During each iteration, we first train the discriminator with objective
\begin{equation}
\begin{aligned}
    \min_{\phi} \sum_{t \geq 1} 
    (\E_{q(\rvx_0)q(\rvx_{t-1}|\rvx_0)q(\rvx_t | \rvx_{t-1})} [ \mathop{\mathrm{F}} (-D_\phi(\rvx_{t-1}, \rvx_t, \rvc, t))] \\
    +\mathbb{E}_{q(\rvx_t)}\mathbb{E}_{p_{\theta}(\rvx_{t-1} | \rvx_{t})} [\mathop{\mathrm{F}} ( D_\phi(\rvx_{t-1}, \rvx_t, \rvc, t))]).
\end{aligned}  
\label{supp_eq:disc_training}
\end{equation}

Then we train the generator with objective
\begin{equation}
\begin{aligned}
\min_{\theta} \sum_{t \geq 1} (
    \mathbb{E}_{q(\rvx_t)}\mathbb{E}_{p_{\theta}(\rvx_{t-1} | \rvx_{t})} [\mathop{\mathrm{F}} ( -D_\phi(\rvx_{t-1}, \rvx_t, \rvc, t))] \\
+ R \cdot \mathcal{L}_{\text{geo}}),
\end{aligned}  
\end{equation}

where $\mathrm{F}(\cdot)$ denotes the $\mathrm{softplus}(\cdot)$ function and $
\mathcal{L}_{\text{geo}} =\mathcal{L}_\text{recon} + \lambda(\mathcal{L}_\text{pos} + \mathcal{L}_\text{vel} +\mathcal{L}_\text{foot})
$, as stated in the main paper.

Similar to \cite{xiao2021tackling} we add an $R_1$ regularization term \cite{mescheder2018training} to the loss term of the discriminator:
\begin{equation}
\frac{\gamma}{2}\E_{q(\rvx_0)q(\rvx_{t-1}|\rvx_0)q(\rvx_t | \rvx_{t-1})} [ \| \nabla_{x_{t-1}} D_{\phi}(\rvx_{t-1}, \rvx_t, \rvc, t) \|^2].
\end{equation}

In this paper, we use $\gamma = 0.02$ for all tasks.

We train our model using the Adam optimizer \cite{kingma2014adam} with cosine learning rate decay~\cite{xiao2021tackling}. The exponential moving average (EMA) is used during the training of the generator. The batch size is $64$ for all tasks. 

The learning rate of the conditional discriminator is  $1.25\times 10^{-4}$. For the generator, we use a learning rate of $3 \times 10^{-5}$ and $2 \times 10^{-5}$ for action-to-motion and text-to-motion tasks, respectively.

\section{{More Experiments}}
\label{sec:supp_more_exp}

\subsection{Comparisons with DDIM Sampling Methods}
\label{sec:comp:few_step}
Moreover, in Tab.~\ref{supp_tab:humanml3d_10_steps}, we compare EMDM with other few-step sampling diffusion models for motion generation~\cite{zhang2022motiondiffuse, mdm2022human, chen2023executing}. To be specific, we show that accelerating sampling by naively reducing the sampling step size using DDIM (10 steps) leads to quality degradation due to the inaccurate approximation of complex data distributions as analyzed in Sec.1 of the main paper. This holds true for both motion diffusion models~\cite{mdm2022human, zhang2022motiondiffuse} or the motion latent diffusion models~\cite{chen2023executing}.

\begin{table}[t]
\large
\centering
\caption{Comparison with motion diffusion models with few-step sampling ({10 sampling steps}) on Text-to-motion. We test on HumanML3D.}
\label{supp_tab:humanml3d_10_steps}
\resizebox{\textwidth}{!}{
\begin{threeparttable}
\setlength{\tabcolsep}{0.15mm}{
\begin{tabular}{@{}lcccccccc@{}}
\toprule
\multirow{2}{*}{Methods} & \multicolumn{3}{c}{R Precision $\uparrow$}              & \multicolumn{1}{c}{\multirow{2}{*}{FID$\downarrow$}} & \multirow{2}{*}{MM Dist$\downarrow$}              & \multirow{2}{*}{Diversity$\rightarrow$}           
& \multirow{2}{*}{MModality$\uparrow$} 
& \multirow{2}{*}{\makecell[c]{Running Time\\(per frame; ms)$\downarrow$}} 
\\ \cmidrule(lr){2-4}
              & \multicolumn{1}{c}{Top 1} & \multicolumn{1}{c}{Top 2} & \multicolumn{1}{c}{Top 3} & \multicolumn{1}{c}{}                     &                          &                            &                            \\ \midrule
Real &
  $0.511^{\pm.003}$ &
  $0.703^{\pm.003}$ &
  $0.797^{\pm.002}$ &
  $0.002^{\pm.000}$ &
  $2.974^{\pm.008}$ &
  $9.503^{\pm.065}$ &
  \multicolumn{1}{c}{-} &
  -
  \\ 
  \midrule
MotionDiffuse &
 $0.040^{\pm.005}$&
 $0.074^{\pm.006}$&
 $0.108^{\pm.008}$&
 $100.780^{\pm.619}$&
 $12.434^{\pm.052}$&
 $10.943^{\pm.106}$&
 $6.650^{\pm.273}$&
$1.426^{\pm.030}$
  \\
MDM&
  $0.076^{\pm.062}$ &
  $0.139^{\pm.004}$ &
  $0.194^{\pm.007}$ &
  $33.232^{\pm.308}$ &
  $7.165^{\pm.048}$&
  $3.440^{\pm.060}$ &
  \textcolor{top2}{\bm{$2.325^{\pm.023}$}} &
  $0.673^{\pm.001}$  
  \\
MLD$\bm{\dag}$ &
  \textcolor{top2}{\bm{${0.480}^{\pm.003}$}} &
  \textcolor{top2}{\bm{${0.670}^{\pm.003}$}} &
  \textcolor{top2}{\bm{${0.769}^{\pm.003}$}} &
  \textcolor{top2}{\bm{${0.397}^{\pm.009}$}} &
  \textcolor{top2}{\bm{${3.199}^{\pm.010}$}} &
  \textcolor{top2}{\bm{$9.923^{\pm.075}$}} &
  \textcolor{top1}{\bm{${2.488}^{\pm.094}$}} &
  \textcolor{top2}{\bm{$0.359^{\pm.002}$}}\\
   \midrule
EMDM (Ours) &  
   \textcolor{top1}{\bm{$0.498^{\pm.007}$}} & 
   \textcolor{top1}{\bm{$0.684^{\pm.006}$}} & 
   \textcolor{top1}{\bm{$0.786^{\pm.006}$}} & 
   \textcolor{top1}{\bm{$0.112^{\pm.019}$}} & 
   \textcolor{top1}{\bm{$3.110^{\pm.027}$}} & 
   \textcolor{top1}{\bm{$9.551^{\pm.078}$}} & 
   $1.641^{\pm.078}$ & 
   \textcolor{top1}{\bm{$0.280^{\pm.002}$}} 
   \\ 
   \bottomrule
\end{tabular}
}
\begin{tablenotes}
   \item  ~\textcolor{top1}{\textbf{$\text{Blue}$}} and \textcolor{top2}{\textbf{$\text{orange}$}} indicate the best and the second best result.\\
     $\dag$~\textbf{Two-stage and non end-to-end approach.}
\end{tablenotes}
\end{threeparttable}
}
\end{table}

\subsection{Comparisons with DDGAN~\cite{xiao2021tackling}}
\label{sec_supp:ablation_DDGAN}
In addition to the DDIM approach, the recent work DDGAN~\cite{xiao2021tackling} proposes another implementation of a few-step sampling for efficient image generation. Next, we compare EMDM with a baseline model that directly combines DDGAN~\cite{xiao2021tackling} and a representative motion diffusion model MDM~\cite{mdm2022human}. Specifically, the baseline approach trains without a condition passed to the discriminator with the weights of geometric loss set to be $0$~($R=0$). The experiment is conducted using HumanML3D datasets for the text-to-motion task. As shown in Tab.~\ref{tab:naiveddgan}, Naive DDGAN produces poor performance in terms of generated motion quality, which is because motion generation typically requires more specific constraints for each frame of the movement.
\begin{table}[t]
        \centering
        \caption{EMDM v.s. DDGAN on HumanML3D.}
\label{tab:naiveddgan}
\centering
\resizebox{\textwidth}{!}{
\begin{threeparttable}
\setlength{\tabcolsep}{1.5mm}{
\begin{tabular}{@{}lccccccc@{}}
\toprule
\multirow{2}{*}{Methods} & \multicolumn{3}{c}{R Precision $\uparrow$}              & \multicolumn{1}{c}{\multirow{2}{*}{FID$\downarrow$}} & \multirow{2}{*}{MM Dist$\downarrow$}              & \multirow{2}{*}{Diversity$\rightarrow$}           
& \multirow{2}{*}{MModality$\uparrow$} 
\\ \cmidrule(lr){2-4}
              & \multicolumn{1}{c}{Top 1} & \multicolumn{1}{c}{Top 2} & \multicolumn{1}{c}{Top 3} & \multicolumn{1}{c}{}                     &                          &                            &                            \\ \midrule
Real &
  $0.511^{\pm.003}$ &
  $0.703^{\pm.003}$ &
  $0.797^{\pm.002}$ &
  $0.002^{\pm.000}$ &
  $2.974^{\pm.008}$ &
  $9.503^{\pm.065}$ &
  \multicolumn{1}{c}{-} 
  \\ 
  \midrule
Naive DDGAN&
$0.072^{\pm.003}$ & 
$0.140^{\pm.004}$ & 
$0.207^{\pm.006}$ & 
$31.085^{\pm.256}$&
$7.389^{\pm.034}$ & 
$5.060^{\pm.059}$ & 
\textcolor{top1}{\bm{$3.155^{\pm.092}$}} \\

EMDM (Ours) &  
   \textcolor{top1}{\bm{$0.498^{\pm.007}$}} & 
   \textcolor{top1}{\bm{$0.684^{\pm.006}$}} & 
   \textcolor{top1}{\bm{$0.786^{\pm.006}$}} & 
   \textcolor{top1}{\bm{$0.112^{\pm.019}$}} & 
   \textcolor{top1}{\bm{$3.110^{\pm.027}$}} & 
   \textcolor{top1}{\bm{$9.551^{\pm.078}$}} & $1.641^{\pm.078}$ \\
   \bottomrule

\end{tabular}
}
\begin{tablenotes}
    \item ~\textcolor{top1}{\textbf{$\text{Blue}$}} indicates the best result. \\
\end{tablenotes}
\end{threeparttable}
}
\end{table}

\subsection{Ablation Study on Conditioning with Geometric Loss.} 
As shown in Tab.~\ref{supp_tab:d_cond}, without providing conditions to the discriminator, the performance in motion quality is slightly worse. This proves the necessity of providing text/action conditions to the discriminator, which is different from naive DDGAN \cite{xiao2021tackling}.

\begin{table}[t]
\footnotesize
\caption{Influence of condition on discriminator. Both models are trained to the same number of epochs. }
\label{supp_tab:d_cond}
\centering
\setlength{\tabcolsep}{0.15mm}
\resizebox{\textwidth}{!}{
\begin{tabular}{@{}lccccccc@{}}

\toprule
\multirow{2}{*}{\makecell[c]{Diffusion\\Steps}} & \multicolumn{3}{c}{R Precision $\uparrow$}              & \multicolumn{1}{c}{\multirow{2}{*}{FID$\downarrow$}} & \multirow{2}{*}{MM Dist$\downarrow$}              & \multirow{2}{*}{Diversity$\rightarrow$}           
& \multirow{2}{*}{MModality$\uparrow$}  
\\ \cmidrule(lr){2-4}
              & \multicolumn{1}{c}{Top 1} & \multicolumn{1}{c}{Top 2} & \multicolumn{1}{c}{Top 3} & \multicolumn{1}{c}{}                     &                          &                                                      \\ \midrule

Real &
  $0.424^{\pm.005}$ &
  $0.649^{\pm.006}$ &
  $0.779^{\pm.006}$ &
  $0.031^{\pm.004}$ &
  $2.788^{\pm.012}$ &
  $11.08^{\pm.097}$ &
  \multicolumn{1}{c}{-}
  \\ \midrule

Without&
  $0.467^{\pm.006}$ &
  $0.666^{\pm.006}$ &
  $0.771^{\pm.006}$ &
  $0.510^{\pm.037}$ &
  $3.209^{\pm.021}$ &
  $10.01^{\pm.072}$ &
  ${2.221}^{\pm.021}$ 
  \\

With (Ours) &
  $0.476^{\pm.005}$ &
  $0.674^{\pm.004}$ &
  $0.779^{\pm.004}$ &
  $0.506^{\pm.031}$ &
  $3.187^{\pm.017}$ &
  $10.03^{\pm.075}$ &
  ${2.235}^{\pm.039}$ 
   \\
   
   \bottomrule
\end{tabular}
}
\end{table}

\subsection{Physical Plausibility.}
As discussed in the limitation section, kinematics-based motion generation methods currently focus more on motion semantics and typically suffer from physical implausibility. We report penetration and skate metrics following~\cite{yuan2023physdiff} in Tab.~\ref{tab:phys}, evaluated on 200 motions, where our motion quality is better than T2M-GPT and comparable with MDM. We agree that injecting physics information can be a promising future direction. 
\begin{table}[t]   
\caption{Comparison on Physical Plausibility.}
\label{tab:phys}
\centering
    \begin{tabular}{ccc}
    \toprule
        Method &  Penetration & Skate \\
         \midrule
         EMDM&  0.094& 1.083\\
         MDM&  0.064& 0.878\\
         T2M-GPT&  0.356&  2.618\\
         \bottomrule
    \end{tabular}
\end{table}

\section{Limitations and Future Works}
\label{sec:supp_failure_cases}
\begin{figure}[H]
\centering
\begin{overpic}[width=0.6\linewidth]{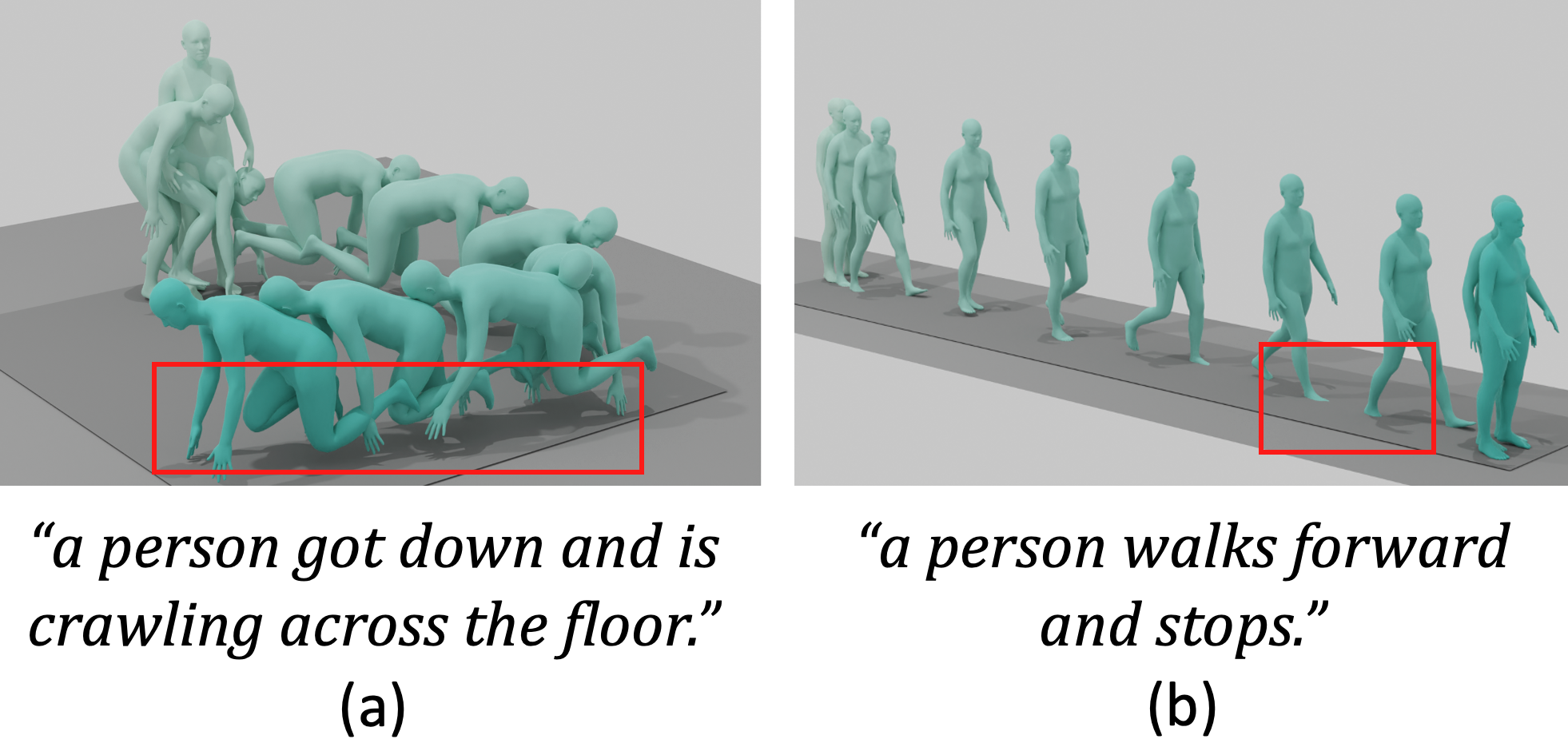}
\end{overpic}
\caption{Motion artifacts: (a) floating and (b) ground penetration in the generated human motion.}
\label{fig_supp:failure_case}
\end{figure}
While EMDM demonstrates promising performance in efficient human motion generation, it lacks physical considerations, leading to issues such as floating and ground penetration; See Fig.~\ref{fig_supp:failure_case}. Integrating physics-based characters shows potential for future improvements. Additionally, although EMDM currently primarily accepts textual inputs, it has the potential to incorporate visual inputs or music sources for online motion synthesis, offering exciting research directions.

\bibliographystyle{splncs04}
\bibliography{main}

\begin{thebibliography}{100}
\providecommand{\url}[1]{\texttt{#1}}
\providecommand{\urlprefix}{URL }
\providecommand{\doi}[1]{https://doi.org/#1}

\bibitem{ahuja2019language2pose}
Ahuja, C., Morency, L.P.: Language2pose: Natural language grounded pose forecasting. In: 2019 International Conference on 3D Vision (3DV). pp. 719--728. IEEE (2019)

\bibitem{alexanderson2023listen}
Alexanderson, S., Nagy, R., Beskow, J., Henter, G.E.: Listen, denoise, action! audio-driven motion synthesis with diffusion models. ACM Transactions on Graphics (TOG)  \textbf{42}(4),  1--20 (2023)

\bibitem{ao2022rhythmic}
Ao, T., Gao, Q., Lou, Y., Chen, B., Liu, L.: Rhythmic gesticulator: Rhythm-aware co-speech gesture synthesis with hierarchical neural embeddings. ACM Transactions on Graphics (TOG)  \textbf{41}(6),  1--19 (2022)

\bibitem{ao2023gesturediffuclip}
Ao, T., Zhang, Z., Liu, L.: Gesturediffuclip: Gesture diffusion model with clip latents. arXiv preprint arXiv:2303.14613  (2023)

\bibitem{cervantes2022implicit}
Cervantes, P., Sekikawa, Y., Sato, I., Shinoda, K.: Implicit neural representations for variable length human motion generation. In: European Conference on Computer Vision. pp. 356--372. Springer (2022)

\bibitem{chen2024taming}
Chen, R., Shi, M., Huang, S., Tan, P., Komura, T., Chen, X.: Taming diffusion probabilistic models for character control. arXiv preprint arXiv:2404.15121  (2024)

\bibitem{chen2023executing}
Chen, X., Jiang, B., Liu, W., Huang, Z., Fu, B., Chen, T., Yu, G.: Executing your commands via motion diffusion in latent space. In: Proceedings of the IEEE/CVF Conference on Computer Vision and Pattern Recognition. pp. 18000--18010 (2023)

\bibitem{chen2022learning}
Chen, X., Su, Z., Yang, L., Cheng, P., Xu, L., Fu, B., Yu, G.: Learning variational motion prior for video-based motion capture. arXiv preprint arXiv:2210.15134  (2022)

\bibitem{chong2020detection}
Chong, E., Clark-Whitney, E., Southerland, A., Stubbs, E., Miller, C., Ajodan, E.L., Silverman, M.R., Lord, C., Rozga, A., Jones, R.M., et~al.: Detection of eye contact with deep neural networks is as accurate as human experts. Nature communications  \textbf{11}(1), ~6386 (2020)

\bibitem{chou2023diffusion}
Chou, G., Bahat, Y., Heide, F.: Diffusion-sdf: Conditional generative modeling of signed distance functions. In: Proceedings of the IEEE/CVF International Conference on Computer Vision. pp. 2262--2272 (2023)

\bibitem{christen2023learning}
Christen, S., Yang, W., P{\'e}rez-D’Arpino, C., Hilliges, O., Fox, D., Chao, Y.W.: Learning human-to-robot handovers from point clouds. In: Proceedings of the IEEE/CVF Conference on Computer Vision and Pattern Recognition. pp. 9654--9664 (2023)

\bibitem{chung2022scaling}
Chung, H.W., Hou, L., Longpre, S., Zoph, B., Tay, Y., Fedus, W., Li, Y., Wang, X., Dehghani, M., Brahma, S., et~al.: Scaling instruction-finetuned language models. arXiv preprint arXiv:2210.11416  (2022)

\bibitem{cong2024laserhuman}
Cong, P., Dou, Z.W., Ren, Y., Yin, W., Cheng, K., Sun, Y., Long, X., Zhu, X., Ma, Y.: Laserhuman: Language-guided scene-aware human motion generation in free environment. arXiv preprint arXiv:2403.13307  (2024)

\bibitem{crawford2022impact}
Crawford, F.W., Jones, S.A., Cartter, M., Dean, S.G., Warren, J.L., Li, Z.R., Barbieri, J., Campbell, J., Kenney, P., Valleau, T., et~al.: Impact of close interpersonal contact on covid-19 incidence: Evidence from 1 year of mobile device data. Science advances  \textbf{8}(1),  eabi5499 (2022)

\bibitem{dabral2023mofusion}
Dabral, R., Mughal, M.H., Golyanik, V., Theobalt, C.: Mofusion: A framework for denoising-diffusion-based motion synthesis. In: Proceedings of the IEEE/CVF Conference on Computer Vision and Pattern Recognition. pp. 9760--9770 (2023)

\bibitem{dou2023c}
Dou, Z., Chen, X., Fan, Q., Komura, T., Wang, W.: C$\cdot$ase: Learning conditional adversarial skill embeddings for physics-based characters. arXiv preprint arXiv:2309.11351  (2023)

\bibitem{dou2023tore}
Dou, Z., Wu, Q., Lin, C., Cao, Z., Wu, Q., Wan, W., Komura, T., Wang, W.: Tore: Token reduction for efficient human mesh recovery with transformer. In: Proceedings of the IEEE/CVF International Conference on Computer Vision. pp. 15143--15155 (2023)

\bibitem{duan2021single}
Duan, Y., Shi, T., Zou, Z., Lin, Y., Qian, Z., Zhang, B., Yuan, Y.: Single-shot motion completion with transformer. arXiv preprint arXiv:2103.00776  (2021)

\bibitem{Guo_2022_CVPR_humanml3d}
Guo, C., Zou, S., Zuo, X., Wang, S., Ji, W., Li, X., Cheng, L.: Generating diverse and natural 3d human motions from text. In: Proceedings of the IEEE/CVF Conference on Computer Vision and Pattern Recognition (CVPR). pp. 5152--5161 (June 2022)

\bibitem{chuan2022tm2t}
Guo, C., Zuo, X., Wang, S., Cheng, L.: Tm2t: Stochastic and tokenized modeling for the reciprocal generation of 3d human motions and texts. In: ECCV (2022)

\bibitem{guo2020action2motion}
Guo, C., Zuo, X., Wang, S., Zou, S., Sun, Q., Deng, A., Gong, M., Cheng, L.: Action2motion: Conditioned generation of 3d human motions. In: Proceedings of the 28th ACM International Conference on Multimedia. pp. 2021--2029 (2020)

\bibitem{guo2023student}
Guo, Y., Dou, Z., Zhang, N., Liu, X., Su, B., Li, Y., Zhang, Y.: Student close contact behavior and covid-19 transmission in china's classrooms. PNAS nexus  \textbf{2}(5),  pgad142 (2023)

\bibitem{harvey2020robust}
Harvey, F.G., Yurick, M., Nowrouzezahrai, D., Pal, C.: Robust motion in-betweening. ACM Transactions on Graphics (TOG)  \textbf{39}(4),  60--1 (2020)

\bibitem{ho2022imagen}
Ho, J., Chan, W., Saharia, C., Whang, J., Gao, R., Gritsenko, A., Kingma, D.P., Poole, B., Norouzi, M., Fleet, D.J., et~al.: Imagen video: High definition video generation with diffusion models. arXiv preprint arXiv:2210.02303  (2022)

\bibitem{ho2020denoising}
Ho, J., Jain, A., Abbeel, P.: Denoising diffusion probabilistic models. Advances in Neural Information Processing Systems  \textbf{33},  6840--6851 (2020)

\bibitem{ho2022classifier}
Ho, J., Salimans, T.: Classifier-free diffusion guidance. arXiv preprint arXiv:2207.12598  (2022)

\bibitem{holden2017phase}
Holden, D., Komura, T., Saito, J.: Phase-functioned neural networks for character control. ACM Transactions on Graphics (TOG)  \textbf{36}(4),  1--13 (2017)

\bibitem{jiang2024motiongpt}
Jiang, B., Chen, X., Liu, W., Yu, J., Yu, G., Chen, T.: Motiongpt: Human motion as a foreign language. Advances in Neural Information Processing Systems  \textbf{36} (2024)

\bibitem{jiang2023drop}
Jiang, Y., Won, J., Ye, Y., Liu, C.K.: Drop: Dynamics responses from human motion prior and projective dynamics. arXiv preprint arXiv:2309.13742  (2023)

\bibitem{karunratanakul2023guided}
Karunratanakul, K., Preechakul, K., Suwajanakorn, S., Tang, S.: Guided motion diffusion for controllable human motion synthesis. In: Proceedings of the IEEE/CVF International Conference on Computer Vision. pp. 2151--2162 (2023)

\bibitem{kim2022flame}
Kim, J., Kim, J., Choi, S.: Flame: Free-form language-based motion synthesis \& editing. arXiv preprint arXiv:2209.00349  (2022)

\bibitem{kingma2014adam}
Kingma, D.P., Ba, J.: Adam: A method for stochastic optimization. arXiv preprint arXiv:1412.6980  (2014)

\bibitem{kong2023priority}
Kong, H., Gong, K., Lian, D., Mi, M.B., Wang, X.: Priority-centric human motion generation in discrete latent space. In: Proceedings of the IEEE/CVF International Conference on Computer Vision. pp. 14806--14816 (2023)

\bibitem{lee2019dancing}
Lee, H.Y., Yang, X., Liu, M.Y., Wang, T.C., Lu, Y.D., Yang, M.H., Kautz, J.: Dancing to music. Advances in neural information processing systems  \textbf{32} (2019)

\bibitem{questenvsim}
Lee, S., Starke, S., Ye, Y., Won, J., Winkler, A.: Questenvsim: Environment-aware simulated motion tracking from sparse sensors. arXiv preprint arXiv:2306.05666  (2023)

\bibitem{lee2023multiact}
Lee, T., Moon, G., Lee, K.M.: Multiact: Long-term 3d human motion generation from multiple action labels. In: Proceedings of the AAAI Conference on Artificial Intelligence. vol.~37, pp. 1231--1239 (2023)

\bibitem{li2022danceformer}
Li, B., Zhao, Y., Zhelun, S., Sheng, L.: Danceformer: Music conditioned 3d dance generation with parametric motion transformer. In: Proceedings of the AAAI Conference on Artificial Intelligence. vol.~36, pp. 1272--1279 (2022)

\bibitem{li2021hybrik}
Li, J., Xu, C., Chen, Z., Bian, S., Yang, L., Lu, C.: Hybrik: A hybrid analytical-neural inverse kinematics solution for 3d human pose and shape estimation. In: Proceedings of the IEEE/CVF conference on computer vision and pattern recognition. pp. 3383--3393 (2021)

\bibitem{li2021ai}
Li, R., Yang, S., Ross, D.A., Kanazawa, A.: Ai choreographer: Music conditioned 3d dance generation with aist++. In: Proceedings of the IEEE/CVF International Conference on Computer Vision. pp. 13401--13412 (2021)

\bibitem{li2023aamdm}
Li, T., Qiao, C., Ren, G., Yin, K., Ha, S.: Aamdm: Accelerated auto-regressive motion diffusion model. arXiv preprint arXiv:2401.06146  (2023)

\bibitem{li2023robust}
Li, Z., Peng, X.B., Abbeel, P., Levine, S., Berseth, G., Sreenath, K.: Robust and versatile bipedal jumping control through reinforcement learning. Robotics: Science and Systems XIX, Daegu, Republic of Korea  (2023)

\bibitem{liao2024vinecs}
Liao, Z., Golyanik, V., Habermann, M., Theobalt, C.: Vinecs: Video-based neural character skinning. In: Proceedings of the IEEE/CVF Conference on Computer Vision and Pattern Recognition. pp. 1377--1387 (2024)

\bibitem{liao2022skeleton}
Liao, Z., Yang, J., Saito, J., Pons-Moll, G., Zhou, Y.: Skeleton-free pose transfer for stylized 3d characters. In: European Conference on Computer Vision. pp. 640--656. Springer (2022)

\bibitem{liu2022close}
Liu, X., Dou, Z., Wang, L., Su, B., Jin, T., Guo, Y., Wei, J., Zhang, N.: Close contact behavior-based covid-19 transmission and interventions in a subway system. Journal of Hazardous Materials  \textbf{436},  129233 (2022)

\bibitem{liu2023syncdreamer}
Liu, Y., Lin, C., Zeng, Z., Long, X., Liu, L., Komura, T., Wang, W.: Syncdreamer: Generating multiview-consistent images from a single-view image. arXiv preprint arXiv:2309.03453  (2023)

\bibitem{long2023wonder3d}
Long, X., Guo, Y.C., Lin, C., Liu, Y., Dou, Z., Liu, L., Ma, Y., Zhang, S.H., Habermann, M., Theobalt, C., et~al.: Wonder3d: Single image to 3d using cross-domain diffusion. arXiv preprint arXiv:2310.15008  (2023)

\bibitem{AMASS_ICCV2019}
Mahmood, N., Ghorbani, N., Troje, N.F., Pons-Moll, G., Black, M.J.: Amass: Archive of motion capture as surface shapes. In: Proceedings of the IEEE/CVF International Conference on Computer Vision (ICCV) (October 2019)

\bibitem{mescheder2018training}
Mescheder, L., Geiger, A., Nowozin, S.: Which training methods for gans do actually converge? In: International conference on machine learning. pp. 3481--3490. PMLR (2018)

\bibitem{muller2023diffrf}
M{\"u}ller, N., Siddiqui, Y., Porzi, L., Bulo, S.R., Kontschieder, P., Nie{\ss}ner, M.: Diffrf: Rendering-guided 3d radiance field diffusion. In: Proceedings of the IEEE/CVF Conference on Computer Vision and Pattern Recognition. pp. 4328--4338 (2023)

\bibitem{pang2023bodyformer}
Pang, K., Qin, D., Fan, Y., Habekost, J., Shiratori, T., Yamagishi, J., Komura, T.: Bodyformer: Semantics-guided 3d body gesture synthesis with transformer. ACM Transactions on Graphics (TOG)  \textbf{42}(4),  1--12 (2023)

\bibitem{vposer_SMPL-X:2019}
Pavlakos, G., Choutas, V., Ghorbani, N., Bolkart, T., Osman, A.A.A., Tzionas, D., Black, M.J.: Expressive body capture: 3d hands, face, and body from a single image. In: Proceedings IEEE Conf. on Computer Vision and Pattern Recognition (CVPR) (2019)

\bibitem{peng2018deepmimic}
Peng, X.B., Abbeel, P., Levine, S., Van~de Panne, M.: Deepmimic: Example-guided deep reinforcement learning of physics-based character skills. ACM Transactions On Graphics (TOG)  \textbf{37}(4),  1--14 (2018)

\bibitem{peng2022ase}
Peng, X.B., Guo, Y., Halper, L., Levine, S., Fidler, S.: Ase: Large-scale reusable adversarial skill embeddings for physically simulated characters. ACM Transactions On Graphics (TOG)  \textbf{41}(4),  1--17 (2022)

\bibitem{2021-TOG-AMP}
Peng, X.B., Ma, Z., Abbeel, P., Levine, S., Kanazawa, A.: Amp: Adversarial motion priors for stylized physics-based character control. ACM Trans. Graph.  \textbf{40}(4) (Jul 2021). \doi{10.1145/3450626.3459670}, \url{http://doi.acm.org/10.1145/3450626.3459670}

\bibitem{petrovich21actor}
Petrovich, M., Black, M.J., Varol, G.: Action-conditioned 3{D} human motion synthesis with transformer {VAE}. In: International Conference on Computer Vision (ICCV) (2021)

\bibitem{petrovich22temos}
Petrovich, M., Black, M.J., Varol, G.: {TEMOS}: Generating diverse human motions from textual descriptions. In: European Conference on Computer Vision ({ECCV}) (2022)

\bibitem{petrovich2022temos}
Petrovich, M., Black, M.J., Varol, G.: Temos: Generating diverse human motions from textual descriptions. In: European Conference on Computer Vision. pp. 480--497. Springer (2022)

\bibitem{pi2023hierarchical}
Pi, H., Peng, S., Yang, M., Zhou, X., Bao, H.: Hierarchical generation of human-object interactions with diffusion probabilistic models. In: Proceedings of the IEEE/CVF International Conference on Computer Vision. pp. 15061--15073 (2023)

\bibitem{Plappert2016kit}
Plappert, M., Mandery, C., Asfour, T.: The kit motion-language dataset. Big Data  \textbf{4}(4),  236--252 (dec 2016). \doi{10.1089/big.2016.0028}, \url{http://dx.doi.org/10.1089/big.2016.0028}

\bibitem{po2023state}
Po, R., Yifan, W., Golyanik, V., Aberman, K., Barron, J.T., Bermano, A.H., Chan, E.R., Dekel, T., Holynski, A., Kanazawa, A., et~al.: State of the art on diffusion models for visual computing. arXiv preprint arXiv:2310.07204  (2023)

\bibitem{raab2022modi}
Raab, S., Leibovitch, I., Li, P., Aberman, K., Sorkine-Hornung, O., Cohen-Or, D.: Modi: Unconditional motion synthesis from diverse data. arXiv preprint arXiv:2206.08010  (2022)

\bibitem{radford2021learning}
Radford, A., Kim, J.W., Hallacy, C., Ramesh, A., Goh, G., Agarwal, S., Sastry, G., Askell, A., Mishkin, P., Clark, J., et~al.: Learning transferable visual models from natural language supervision. In: International Conference on Machine Learning. pp. 8748--8763. PMLR (2021)

\bibitem{raffel2020exploring}
Raffel, C., Shazeer, N., Roberts, A., Lee, K., Narang, S., Matena, M., Zhou, Y., Li, W., Liu, P.J.: Exploring the limits of transfer learning with a unified text-to-text transformer. The Journal of Machine Learning Research  \textbf{21}(1),  5485--5551 (2020)

\bibitem{rempe2021humor}
Rempe, D., Birdal, T., Hertzmann, A., Yang, J., Sridhar, S., Guibas, L.J.: Humor: 3d human motion model for robust pose estimation. In: International Conference on Computer Vision (ICCV) (2021)

\bibitem{stable_diffusion}
Rombach, R., Blattmann, A., Lorenz, D., Esser, P., Ommer, B.: High-resolution image synthesis with latent diffusion models. In: Proceedings of the IEEE Conference on Computer Vision and Pattern Recognition (CVPR) (2022), \url{https://github.com/CompVis/latent-diffusionhttps://arxiv.org/abs/2112.10752}

\bibitem{shi2020motionet}
Shi, M., Aberman, K., Aristidou, A., Komura, T., Lischinski, D., Cohen-Or, D., Chen, B.: Motionet: 3d human motion reconstruction from monocular video with skeleton consistency. ACM Transactions on Graphics (TOG)  \textbf{40}(1),  1--15 (2020)

\bibitem{shi2023phasemp}
Shi, M., Starke, S., Ye, Y., Komura, T., Won, J.: Phasemp: Robust 3d pose estimation via phase-conditioned human motion prior. In: Proceedings of the IEEE/CVF International Conference on Computer Vision. pp. 14725--14737 (2023)

\bibitem{shi2023controllable}
Shi, Y., Wang, J., Jiang, X., Dai, B.: Controllable motion diffusion model. arXiv preprint arXiv:2306.00416  (2023)

\bibitem{smith2023learning}
Smith, L., Kew, J.C., Li, T., Luu, L., Peng, X.B., Ha, S., Tan, J., Levine, S.: Learning and adapting agile locomotion skills by transferring experience. arXiv preprint arXiv:2304.09834  (2023)

\bibitem{sohl2015deep}
Sohl-Dickstein, J., Weiss, E., Maheswaranathan, N., Ganguli, S.: Deep unsupervised learning using nonequilibrium thermodynamics. In: International Conference on Machine Learning. pp. 2256--2265. PMLR (2015)

\bibitem{song2020denoising}
Song, J., Meng, C., Ermon, S.: Denoising diffusion implicit models. arXiv preprint arXiv:2010.02502  (2020)

\bibitem{starke2022deepphase}
Starke, S., Mason, I., Komura, T.: Deepphase: periodic autoencoders for learning motion phase manifolds. ACM Transactions on Graphics (TOG)  \textbf{41}(4),  1--13 (2022)

\bibitem{starke2019neural}
Starke, S., Zhang, H., Komura, T., Saito, J.: Neural state machine for character-scene interactions. ACM Trans. Graph.  \textbf{38}(6),  209--1 (2019)

\bibitem{starke2021neural}
Starke, S., Zhao, Y., Zinno, F., Komura, T.: Neural animation layering for synthesizing martial arts movements. ACM Transactions on Graphics (TOG)  \textbf{40}(4),  1--16 (2021)

\bibitem{sun2024aios}
Sun, Q., Wang, Y., Zeng, A., Yin, W., Wei, C., Wang, W., Mei, H., Leung, C.S., Liu, Z., Yang, L., et~al.: Aios: All-in-one-stage expressive human pose and shape estimation. In: Proceedings of the IEEE/CVF Conference on Computer Vision and Pattern Recognition. pp. 1834--1843 (2024)

\bibitem{tessler2023calm}
Tessler, C., Kasten, Y., Guo, Y., Mannor, S., Chechik, G., Peng, X.B.: Calm: Conditional adversarial latent models for directable virtual characters. In: ACM SIGGRAPH 2023 Conference Proceedings. pp.~1--9 (2023)

\bibitem{tevet2022motionclip}
Tevet, G., Gordon, B., Hertz, A., Bermano, A.H., Cohen-Or, D.: Motionclip: Exposing human motion generation to clip space. arXiv preprint arXiv:2203.08063  (2022)

\bibitem{mdm2022human}
Tevet, G., Raab, S., Gordon, B., Shafir, Y., Bermano, A.H., Cohen-Or, D.: Human motion diffusion model. arXiv preprint arXiv:2209.14916  (2022)

\bibitem{touvron2023llama}
Touvron, H., Lavril, T., Izacard, G., Martinet, X., Lachaux, M.A., Lacroix, T., Rozi{\`e}re, B., Goyal, N., Hambro, E., Azhar, F., et~al.: Llama: Open and efficient foundation language models. arXiv preprint arXiv:2302.13971  (2023)

\bibitem{voas2309best}
Voas, J.: What is the best automated metric for text to motion generation? arxiv 2023. arXiv preprint arXiv:2309.10248

\bibitem{wan2023tlcontrol}
Wan, W., Dou, Z., Komura, T., Wang, W., Jayaraman, D., Liu, L.: Tlcontrol: Trajectory and language control for human motion synthesis. arXiv preprint arXiv:2311.17135  (2023)

\bibitem{wan2023diffusionphase}
Wan, W., Huang, Y., Wu, S., Komura, T., Wang, W., Jayaraman, D., Liu, L.: Diffusionphase: Motion diffusion in frequency domain. arXiv preprint arXiv:2312.04036  (2023)

\bibitem{wan2022learn}
Wan, W., Yang, L., Liu, L., Zhang, Z., Jia, R., Choi, Y.K., Pan, J., Theobalt, C., Komura, T., Wang, W.: Learn to predict how humans manipulate large-sized objects from interactive motions. IEEE Robotics and Automation Letters  \textbf{7}(2),  4702--4709 (2022)

\bibitem{wan2022}
Wan, W., Yang, L., Liu, L., Zhang, Z., Jia, R., Choi, Y.K., Pan, J., Theobalt, C., Komura, T., Wang, W.: Learn to predict how humans manipulate large-sized objects from interactive motions. IEEE Robotics and Automation Letters  \textbf{7}(2),  4702--4709 (2022). \doi{10.1109/LRA.2022.3151614}

\bibitem{wang2023zolly}
Wang, W., Ge, Y., Mei, H., Cai, Z., Sun, Q., Wang, Y., Shen, C., Yang, L., Komura, T.: Zolly: Zoom focal length correctly for perspective-distorted human mesh reconstruction. arXiv preprint arXiv:2303.13796  (2023)

\bibitem{questsim}
Winkler, A., Won, J., Ye, Y.: Questsim: Human motion tracking from sparse sensors with simulated avatars. In: SIGGRAPH Asia 2022 Conference Papers. pp.~1--8 (2022)

\bibitem{xiao2021tackling}
Xiao, Z., Kreis, K., Vahdat, A.: Tackling the generative learning trilemma with denoising diffusion gans. arXiv preprint arXiv:2112.07804  (2021)

\bibitem{xie2023omnicontrol}
Xie, Y., Jampani, V., Zhong, L., Sun, D., Jiang, H.: Omnicontrol: Control any joint at any time for human motion generation. arXiv preprint arXiv:2310.08580  (2023)

\bibitem{xu2023actformer}
Xu, L., Song, Z., Wang, D., Su, J., Fang, Z., Ding, C., Gan, W., Yan, Y., Jin, X., Yang, X., et~al.: Actformer: A gan-based transformer towards general action-conditioned 3d human motion generation. In: Proceedings of the IEEE/CVF International Conference on Computer Vision. pp. 2228--2238 (2023)

\bibitem{yamane2013synthesizing}
Yamane, K., Revfi, M., Asfour, T.: Synthesizing object receiving motions of humanoid robots with human motion database. In: 2013 IEEE International Conference on Robotics and Automation. pp. 1629--1636. IEEE (2013)

\bibitem{yan2019convolutional}
Yan, S., Li, Z., Xiong, Y., Yan, H., Lin, D.: Convolutional sequence generation for skeleton-based action synthesis. In: Proceedings of the IEEE/CVF International Conference on Computer Vision. pp. 4394--4402 (2019)

\bibitem{yang2023analysis}
Yang, X., Dou, Z., Ding, Y., Su, B., Qian, H., Zhang, N.: Analysis of sars-cov-2 transmission in airports based on real human close contact behaviors. Journal of Building Engineering p. 108299 (2023)

\bibitem{neural3points}
Ye, Y., Liu, L., Hu, L., Xia, S.: Neural3points: Learning to generate physically realistic full-body motion for virtual reality users. In: Computer Graphics Forum. vol.~41, pp. 183--194. Wiley Online Library (2022)

\bibitem{yu2023surf}
Yu, Z., Dou, Z., Long, X., Lin, C., Li, Z., Liu, Y., M{\"u}ller, N., Komura, T., Habermann, M., Theobalt, C., et~al.: Surf-d: High-quality surface generation for arbitrary topologies using diffusion models. arXiv preprint arXiv:2311.17050  (2023)

\bibitem{yuan2023physdiff}
Yuan, Y., Song, J., Iqbal, U., Vahdat, A., Kautz, J.: Physdiff: Physics-guided human motion diffusion model. In: Proceedings of the IEEE/CVF International Conference on Computer Vision. pp. 16010--16021 (2023)

\bibitem{zhang2023learning}
Zhang, H., Yuan, Y., Makoviychuk, V., Guo, Y., Fidler, S., Peng, X.B., Fatahalian, K.: Learning physically simulated tennis skills from broadcast videos. ACM Transactions On Graphics (TOG)  \textbf{42}(4),  1--14 (2023)

\bibitem{zhang2023t2m}
Zhang, J., Zhang, Y., Cun, X., Huang, S., Zhang, Y., Zhao, H., Lu, H., Shen, X.: T2m-gpt: Generating human motion from textual descriptions with discrete representations. arXiv preprint arXiv:2301.06052  (2023)

\bibitem{zhang2023tapmo}
Zhang, J., Huang, S., Tu, Z., Chen, X., Zhan, X., Yu, G., Shan, Y.: Tapmo: Shape-aware motion generation of skeleton-free characters. arXiv preprint arXiv:2310.12678  (2023)

\bibitem{zhang2023skinned}
Zhang, J., Weng, J., Kang, D., Zhao, F., Huang, S., Zhe, X., Bao, L., Shan, Y., Wang, J., Tu, Z.: Skinned motion retargeting with residual perception of motion semantics \& geometry. In: Proceedings of the IEEE/CVF Conference on Computer Vision and Pattern Recognition. pp. 13864--13872 (2023)

\bibitem{zhang2022motiondiffuse}
Zhang, M., Cai, Z., Pan, L., Hong, F., Guo, X., Yang, L., Liu, Z.: Motiondiffuse: Text-driven human motion generation with diffusion model. arXiv preprint arXiv:2208.15001  (2022)

\bibitem{zhang2023remodiffuse}
Zhang, M., Guo, X., Pan, L., Cai, Z., Hong, F., Li, H., Yang, L., Liu, Z.: Remodiffuse: Retrieval-augmented motion diffusion model. arXiv preprint arXiv:2304.01116  (2023)

\bibitem{zhang2023close}
Zhang, N., Liu, L., Dou, Z., Liu, X., Yang, X., Miao, D., Guo, Y., Gu, S., Li, Y., Qian, H., et~al.: Close contact behaviors of university and school students in 10 indoor environments. Journal of Hazardous Materials  \textbf{458},  132069 (2023)

\bibitem{zhang2023popularization}
Zhang, N., Liu, X., Gao, S., Su, B., Dou, Z.: Popularization of high-speed railway reduces the infection risk via close contact route during journey. Sustainable Cities and Society  \textbf{99},  104979 (2023)

\bibitem{zhang2020perpetual}
Zhang, Y., Black, M.J., Tang, S.: Perpetual motion: Generating unbounded human motion. arXiv preprint arXiv:2007.13886  (2020)

\bibitem{zhang2021we}
Zhang, Y., Black, M.J., Tang, S.: We are more than our joints: Predicting how 3d bodies move. In: Proceedings of the IEEE/CVF Conference on Computer Vision and Pattern Recognition. pp. 3372--3382 (2021)

\bibitem{zhang2023motiongpt}
Zhang, Y., Huang, D., Liu, B., Tang, S., Lu, Y., Chen, L., Bai, L., Chu, Q., Yu, N., Ouyang, W.: Motiongpt: Finetuned llms are general-purpose motion generators. arXiv preprint arXiv:2306.10900  (2023)

\bibitem{zhao2020bayesian}
Zhao, R., Su, H., Ji, Q.: Bayesian adversarial human motion synthesis. In: Proceedings of the IEEE/CVF Conference on Computer Vision and Pattern Recognition. pp. 6225--6234 (2020)

\bibitem{zhu2023taming}
Zhu, L., Liu, X., Liu, X., Qian, R., Liu, Z., Yu, L.: Taming diffusion models for audio-driven co-speech gesture generation. In: Proceedings of the IEEE/CVF Conference on Computer Vision and Pattern Recognition. pp. 10544--10553 (2023)

\end{thebibliography}

\end{document}